\def\bb{{\bf b}}

\def\W{{\bf W}}

\def\X{{\bf X}}

\def\X{{\bf X}}
\def\x{{\bf x}}

\def\0{{\bf 0}}
\def\1{{\bf 1}}

\def\ME{{\mathbb E}}

\def\BR{{\mathbb R}}

\def\muu{\mbox{\boldmath$\mu$\unboldmath}}
\def\Si{\mbox{\boldmath$\Sigma$\unboldmath}}

\def\diag{\mathrm{diag}}

\def\diag{\mathrm{diag}}

\newcommand{\col}[2] {#1_{\cdot #2}}

\documentclass{article} % For LaTeX2e
\usepackage[table]{xcolor}
\usepackage{iclr2017_conference,times}
\usepackage{times}
\usepackage{helvet}
\usepackage{courier}
\usepackage{graphicx}
\usepackage{amsmath,amssymb}
\usepackage{amsmath,amssymb}
\usepackage{url}

\usepackage{subfigure}
\usepackage{algorithm}
\usepackage{algorithmic}
\usepackage{multirow}

\graphicspath{{fig/}}

\title{Revisiting Batch Normalization For \\Practical Domain Adaptation}

\author{Yanghao Li$^\dagger$, Naiyan Wang$^\ddagger$, Jianping Shi$^\diamond$, Jiaying Liu$^\dagger$, Xiaodi Hou$^\ddagger$\\
$^\dagger$ Institute of Computer Science and Technology, Peking University\\
$^\ddagger$ TuSimple ~~~
$^\diamond$ SenseTime\\
{\tt\small lyttonhao@pku.edu.cn}~~~{\tt\small winsty@gmail.com}~~{\tt\small  shijianping5000@gmail.com}\\
{\tt\small liujiaying@pku.edu.cn}~~
{\tt\small xiaodi.hou@gmail.com}
}

\begin{document}

\maketitle
\rowcolors{2}{white}{gray!25}

\begin{abstract}
Deep neural networks (DNN) have shown unprecedented success in various computer vision applications such as image classification and object detection. However, it is still a common annoyance during the training phase, that one has to prepare at least thousands of labeled images to fine-tune a network to a specific domain. %However, it is still a common (yet inconvenient) practice to prepare at least tens of thousands of labeled images to fine-tune a network on every task before the model is ready to use. 
Recent study~\citep{deeper_bias} shows that a DNN has strong dependency towards the training dataset, and the learned features cannot be easily transferred to a different but relevant task without fine-tuning. In this paper, we propose a simple yet powerful remedy, called \emph{Adaptive Batch Normalization} (AdaBN) to increase the generalization ability of a DNN. %, based on the well-known Batch Normalization (BN) technique~\citep{bn} which has become a standard component in modern deep learning.
By modulating the statistics in all Batch Normalization layers across the network, our approach achieves deep adaptation effect for domain adaptation tasks. In contrary to other deep learning domain adaptation methods, our method does not require additional components, and is parameter-free. It archives state-of-the-art performance despite its surprising simplicity. Furthermore, we demonstrate that our method is complementary with other existing methods.  Combining AdaBN with existing domain adaptation treatments may further improve model performance.
\end{abstract}

\begin{section}{Introduction}

Training a DNN for a new image recognition task is expensive. It requires a large amount of labeled training images that are not easy to obtain. One common practice is to use %a training set from a different source. For instance, one can borrow training data from an existing dataset, or query images from search engines and then label them using Amazon Mechanical Turk. These approaches usually suffer from inferior performance due to dataset discrepancies, or ``dataset bias'', because 1) the distributions of the source domains (third party datasets or Internet images) are often different from the target domain (testing images); and 2) DNN is particularly good at capturing dataset bias in its internal representation~\citep{unbiased}, which eventually leads to overfitting.
labeled data from other related source such as a different public dataset, or harvesting images by keywords from a search engine. Because 1) the distributions of the source domains (third party datasets or Internet images) are often different from the target domain (testing images); and 2) DNN is particularly good at capturing dataset bias in its internal representation~\citep{unbiased}, which eventually leads to overfitting, imperfectly paired training and testing sets usually leads to inferior performance. 

%The big data have greatly facilitated the development of computer vision area in recent years. For instance, one may easily build a huge dataset by crawling the public data in the Internet. However, the trained model on such data can hardly directly applied due to the fact that the distribution of the data from Internet (source domain) are usually different from those in real scenario (target domain). This is widely recognized as the ``dataset bias'' issue ~\citep{unbiased} which is ubiquitous in almost all datasets. This issue poses a great challenge for the generalization ability of the algorithm, which limits the practical use of a multimedia system.p

Known as domain adaptation, the effort to bridge the gap between training and testing data distributions has been discussed several times under the context of deep learning~\citep{ddc,dan,joint,revgrad}. To make the connection between the domain of training and the domain of testing, most of these methods require additional optimization steps and extra parameters. Such additional  computational burden could greatly complicate the training of a DNN which is already intimidating enough for most people.

%To address such challenge, domain adaptation algorithms~\citep{beijbom2012domain,patel2015visual} are developed. In the classical unsupervised setting of domain adaptation, we are given the data samples in both source domain and target domain, but only the labels in source domain are available. The goal is to make use of the unlabeled data in target domain to improve the performance.
%
%At the same time, deep convolutional neural network has liberated its power in various applications in computer vision and multimedia, such as image classification~\citep{alexnet}, action recognition~\citep{twostream} and video event detection~\citep{devnet,dismed}. There are several works that try to cope with the domain adaptation issues in the context of DNN~\citep{ddc,dan,joint,revgrad}. However, most of these methods introduce additional components, such as regularization between source domain and target domain using maximum mean discrepancy(MMD), which slows down the training process of DNNs significantly. This makes those methods can hardly work on large-scale dataset, thus undesirable for practical use.

\begin{figure}[t]
\center{
\includegraphics[width=0.7\linewidth]{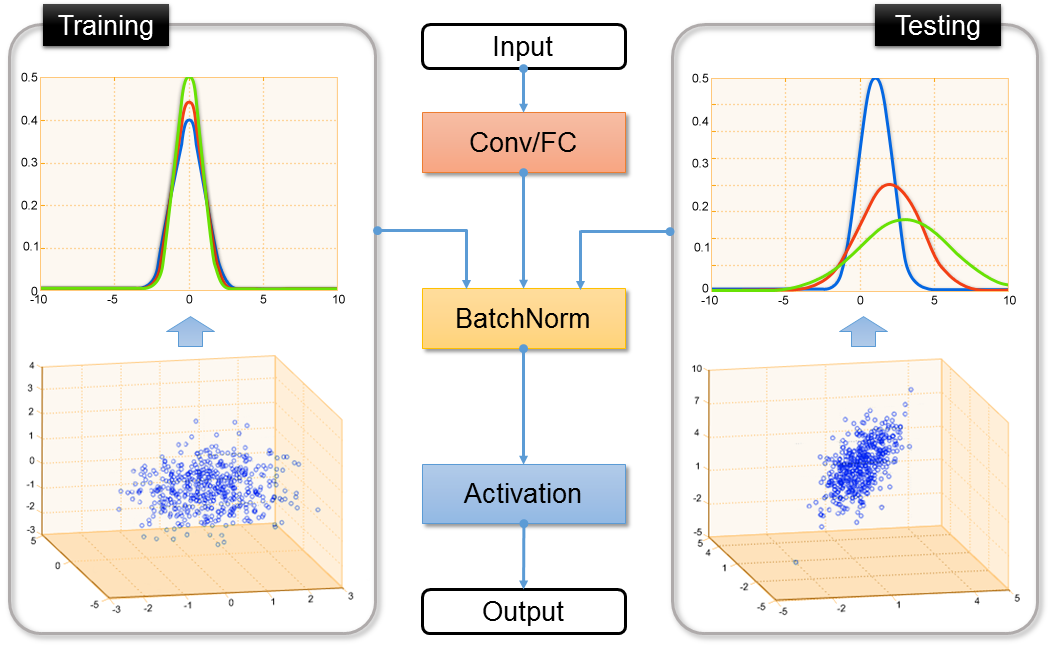}
\caption{Illustration of the proposed method. %In training and testing, we use separate statistics of the output of each convolution or fully connected layer in source domain and target domain for the batch normalization layer. 
For each convolutional or fully connected layer, we use different bias/variance terms to perform batch normalization for the training domain and the test domain. The domain specific normalization mitigates the domain shift issue.}\label{fig:teaser}}
\end{figure}

In this paper, we propose a simple yet effective approach called \emph{AdaBN} for batch normalized DNN domain adaptation. % Our observation suggests a dissociation between bias and variance terms in a batch-normalized DNN. 
We hypothesize that the label related knowledge is stored in the weight matrix of each layer, whereas domain related knowledge is represented by the statistics of the Batch Normalization (BN)~\citep{bn} layer. Therefore, we can easily transfer the trained model to a new domain by modulating the statistics in the BN layer. This approach is straightforward to implement, has zero parameter to tune, and requires minimal computational resources.  Moreover, our AdaBN is ready to be extended to more sophisticated scenarios such as multi-source domain adaptation and semi-supervised settings. Fig.~\ref{fig:teaser} illustrates the flowchart of AdaBN. To summarize, our contributions are as follows:
	
%In this paper, we propose a simple yet effective approach for domain adaptation.  We exploit the intrinsic property of batch normalized DNNs. We conjecture that in a batch normalized CNN, the label related knowledge are learned in the weight of each layer, while the domain related information reside in the statistics of Batch Normalization(BN) layer. Thus separating the statistics in source domain and target domain will help reduce the domain bias. Our method does not add extra components to existing DNNs and is parameter-free, yet it still can yield state-of-the-art performances. It only needs an additional pass of the data in target domain, and can be easily implemented in any deep learning framework. Moreover, it is easy to extend to more sophisticated cases such as multi-source setting and (semi)-supervised settings. An illustration of our proposed method is in Fig.~\ref{fig:teaser}. To summarize, our contribution lists as follows:

\begin{enumerate}
\item We propose a novel domain adaptation technique called Adaptive Batch Normalization (AdaBN). We show that AdaBN can naturally dissociate bias and variance of a dataset, which is ideal for domain adaptation tasks.
\item We validate the effectiveness of our approach on standard benchmarks for both single source and multi-source domain adaptation. Our method outperforms the state-of-the-art methods.
\item We conduct experiments on the cloud detection for remote sensing images to further demonstrate the effectiveness of our approach in practical use. 
\end{enumerate}

\end{section}

\begin{section}{Related Work}
Domain transfer in visual recognition tasks has gained increasing attention in recent literature~\citep{beijbom2012domain,patel2015visual}. Often referred to as \emph{covariate shift}~\citep{shimodaira2000improving} or \emph{dataset bias}~\citep{unbiased}, this problem poses a great challenge to the generalization ability of a learned model. One key component of domain transfer is to model the difference between source and target distributions. In~\citet{khosla2012undoing}, the authors assign each dataset with an explicit bias vector, and train one discriminative model to handle multiple classification problems with different bias terms. A more explicit way to compute dataset difference is based on Maximum Mean Discrepancy (MMD)~\citep{mmd}. This approach  projects each data sample into a Reproducing Kernel Hilbert Space, and then computes the difference of sample means. To reduce dataset discrepancies, many methods are proposed, including sample selections~\citep{huang2006correcting,landmark}, explicit projection learning~\citep{tca,gopalan2011domain,dip} and principal axes alignment~\citep{sa,gfk,aljundi2015landmarks}.

All of these methods face the same challenge of %devising an effective domain transfer function in high-dimensional non-linear space.
constructing the domain transfer function -- a high-dimensional non-linear function. Due to computational constraints, most of the proposed transfer functions are in the category of simple shallow projections, which are typically composed of kernel transformations and linear mapping functions.

%The concept of dataset bias or domain discrepancy in visual recognition tasks is first elaborated in~\citep{unbiased}. It suggests that all existing datasets are biased against the real visual world. Domain adaptation is an essential step before applying one trained model to real scenario. The key to domain adaption is to measure and reduce the distance of distributions in different domains. A common choice for measurement is the maximum mean discrepancy(MMD)~\citep{mmd}. It uses a non-linear mapping to map each data sample to Reproducing Kernel Hilbert Space, and then compute the difference of sample means. To reduce the discrepancy under different criteria, many methods are attempted, including sample selections~\citep{huang2006correcting,landmark}, explicit projection learning~\citep{tca,gopalan2011domain,dip} and principal axes alignment~\citep{sa,gfk}. However, the performance of these models are all restricted by the simplicity of the transformation, since real domain shift in high dimensional space is highly non-linear. For complete surveys of these methods in visual recognition tasks, we refer the readers to~\citep{beijbom2012domain,patel2015visual}.

In the field of deep learning, feature transferability across different domains is a tantalizing yet generally unsolved topic~\citep{transferable,deeper_bias}. To transfer the learned representations to a new dataset, pre-training plus fine-tuning~\citep{decaf} have become \textit{de facto} procedures.  However, adaptation by fine-tuning is far from perfect. It requires a considerable amount of labeled data from the target domain, and non-negligible computational resources to re-train the whole network.

%Recently, convolutional neural networks(CNN) based feature representation has demonstrated its transferability~\citep{transferable} across different datasets and applications~\citep{decaf}. Nevertheless, it also suffers from the dataset bias issue as pointed out in ~\citep{deeper_bias}. To transfer the features across tasks and dataset, pre-training and fine-tuning have become \textit{de facto} procedures in deep learning pipeline. However, adaptation by fine-tuning is far from perfect. It requires a number of labeled data in target domain as supervision, otherwise the CNN is prone to be overfitting due to its high complexity. In the extreme case, unsupervised domain adaptation provides no labels in target domain. Besides, adapting the distribution shift by changing the weight may not be the most essential method.

A series of progress has been made in DNN to facilitate domain transfer. Early works of domain adaptation either focus on reordering fine-tuning samples~\citep{dlid}, or regularizing MMD~\citep{mmd} in a shallow network~\citep{ae_adaptation}. It is only until recently that the problem is directly attacked under the setting of classification of unlabeled target domain using modern convolutional neural network (CNN) architecture. DDC~\citep{ddc} used the classical MMD loss to regularize the representation in the last layer of CNN. DAN~\citep{dan} further extended the method to multiple kernel MMD and multiple layer adaptation. Besides adapting features using MMD, RTN~\citep{long2016unsupervised} also added a gated residual layer for classifier adaptation. RevGrad~\citep{revgrad} devised a gradient reversal layer to %reverse the gradient from the domain classification loss. %that helps to distinguish the domains of each data sample.
%This layer efficiently anonymized the domain information hidden in the CNN features. %Tzeng \etal~\citep{joint} proposed to simultaneously transfer task correlations and maximize domain confusion for (semi)-supervised domain adaptation.
compensate the back-propagated gradients that are domain specific.
Recently,  by explicitly modeling both private and shared components of the domain representations in the network, \citet{bousmalis2016domain} proposed a Domain Separation Network to extract better domain-invariant features.

%Some early works on applying domain adaptation to visual recognition tasks in DNN include~\citep{ae_adaptation,dlid}: ~\citep{dlid} tries to interpolate the representations between the source domain and target domain in DNN, while ~\citep{ae_adaptation} builds the model upon the traditional SURF feature~\citep{surf}. It trains another two-layer denoising autoencoder~\citep{sdae} regularized by MMD to reduce the domain discrepancy. However, due to their relatively shallow models (no more than two stages), their performances still show limited improvements over the conventional methods. After that, several domain adaptation methods are tightly incorporated into CNN: Tzeng \etal~\citep{ddc} used the classical MMD loss~\citep{mmd} to regularize the representation in the last layer of CNN. Long \etal~\citep{dan} further extend it to multiple kernel MMD and multiple layer adaptation cases. Ganin \etal~\citep{revgrad} devised a gradient reverse layer to reverse the gradient that helps to distinguish the domains of each data sample. This layer efficiently anonymizes the domain information hidden in the CNN features. Recently, Tzeng \etal~\citep{joint} proposed to simultaneously transfer task correlations and maximize domain confusion for (semi)-supervised domain adaptation.

Another related work is CORAL~\citep{coral}. This model focuses on the last layer of CNN. CORAL whitens the data in source domain, and then re-correlates the source domain features to target domain. This operation aligns the second order statistics of source domain and target domain distributions. Surprisingly, such simple approach yields state-of-the-arts results in various text classification and visual recognition tasks. Recently, Deep CORAL~\citep{sun2016deep} also extends the method into DNN by incorporating a CORAL loss.

%Another related work is~\citep{coral}. It shares the same spirit of our methods, but is only applied on the features produced by last layer of CNN. It first whitens both the data in source domain and target domain, and then re-correlate the source domain features to target domain. This operation aligns the second order statistics of source domain and target domain distributions. Surprisingly, this simple approach yields state-of-the-art results in various text classification and visual recognition tasks.

\subsection{Batch Normalization}\label{sec:bn}
In this section, we briefly review Batch Normalization (BN)~\citep{bn} which is closely related to our AdaBN. The BN layer is originally designed to alleviate the issue of internal covariate shifting -- a common problem while training a very deep neural network. It first standardizes each feature in a mini-batch, and then learns a common slope and bias for each mini-batch. Formally, given the input to a BN layer $\X \in \BR^{n \times p}$, where $n$ denotes the batch size, and $p$ is the feature dimension, BN layer transforms a feature $j \in \{1 \ldots p\}$ into:

\begin{equation}
	\begin{aligned}
	\hat{x}_j &= \frac{x_j - \ME[\col{\X}{j}]}{\sqrt{\text{Var}[\col{\X}{j}]}}, \\
	y_j &= \gamma_j \hat{x}_j + \beta_j,
	\end{aligned}
\end{equation}
%\begin{equation}
%	\begin{aligned}
%	\hat{\col{\x}{j}} &= \frac{\col{\x}{j} - \ME[\x]}{\sqrt{\text{Var}[\col{\x}{j}]}} \\
%	\col{\y}{j} &= \gamma_j \hat{\col{\x}{j}} + \beta_j,
%	\end{aligned}
%\end{equation}
where $x_j$ and $y_j$ are the input/output scalars of one neuron response in one data sample; $\col{\X}{j}$ denotes the $j^{th}$ column of the input data; and $\gamma_j$ and $\beta_j$ are parameters to be learned. This transformation guarantees that the input distribution of each layer remains unchanged across different mini-batches. For Stochastic Gradient Descent (SGD) optimization, a stable input distribution could greatly facilitate model convergence, leading to much faster training speed for CNN. Moreover, if training data are shuffled at each epoch, the same %training sample is transformed, or augmented differently in each epoch. This property acts as an additional regularization to combat against overfitting. In the testing phase, the global statistics instead of the statistics from one mini-batch are used to stabilize the testing results.
training sample will be applied with different transformations, or in other words, more comprehensively augmented throughout the training. During the testing phase, the global statistics of all training samples is used to normalize every mini-batch of test data.

Extensive experiments have shown that Batch Normalization significantly reduces the number of iteration to converge, and improves the final performance at the same time. BN layer has become a standard component in recent top-performing CNN architectures, such as deep residual network~\citep{resnet}, and Inception V3~\citep{inception_v3}.

\end{section}

\begin{section}{The Model}
In Sec.~\ref{sec:observation}, we first analyze the domain shift in deep neural network, and reveal two key observations. Then in Sec.~\ref{sec:method}, we introduce our Adaptive Batch Normalization (AdaBN) method based on these observations. Finally, we analyze our method in-depth in Sec.~\ref{sec:discuss}.

\begin{subsection}{A Pilot Experiment}\label{sec:observation}
Although the Batch Normalization (BN) technique is originally proposed to help SGD optimization, its core idea is to align the distribution of  training data. From this perspective, it is interesting to examine the BN parameters (batch-wise mean and variance) over different dataset at different layers of the network.

%Batch Normalization (BN) layer is used to transform the sample distribution for optimization efficiency. Meanwhile the sample mean and variance are usually related to data distribution, which is the key factor for domain adaptation. This raises an important question, is such clue can be used for domain adaptation?

In this pilot experiment, we use MXNet implementation~\citep{mxnet} of the Inception-BN model~\citep{bn} pre-trained on ImageNet classification task~\citep{imagenet} as our baseline DNN model. Our image data are drawn from~\citep{bing-caltech}, which contains the same classes of images from both Caltech-256 dataset~\citep{griffin2007caltech} and Bing image search results. For each mini-batch sampled from one dataset, we concatenate the mean and variance of all neurons from one layer  to form a feature vector.  Using linear SVM, we can almost perfectly classify whether the mini-batch feature vector is from Caltech-256 or Bing dataset. Fig.~\ref{fig:bn_visualize} visualizes the distributions of mini-batch feature vectors from two datasets in 2D. It is clear that  BN statistics from different domains are separated into clusters.

%To validate this hypothesis, we conduct the following experiments: First, we starts from a pre-trained CNN model on ImageNet classification task. Specifically, we adopt the reproduction of Inception-BN model\footnote{\url{http://data.dmlc.ml/mxnet/models/imagenet/}} by MXNet~\citep{mxnet}. We extract the BN statistics of each mini-batch using the data samples from two domains (Caltech-Bing dataset. More details are provided in Sec.~\ref{sec:exp}). Inspired by the ``guess the dataset'' game in~\citep{unbiased}, we then train a linear SVM to predict which domain the BN statistics are from. We simply use the concatenation of mean and variance of each feature in one layer as features. On the held out data, it archives both 100\% accuracy on a shallow layer(the BN layer before the second convolution layer) and a deep layer(the last BN layer of the last Inception module). We find that even a simple linear SVM could distinguish these two domains. The t-SNE~\citep{tsne} visualization in Fig.~\ref{fig:bn_visualize} also confirms this surprising results. The BN statistics from different domains are tightly clustered even reducing their dimensions to 2. These observations are illuminating. It reveals two crucial properties of domain shift within DNN:

\begin{figure}[htb]
\center{
\subfigure[Shallow layer distributions] {\includegraphics[width=0.35\linewidth]{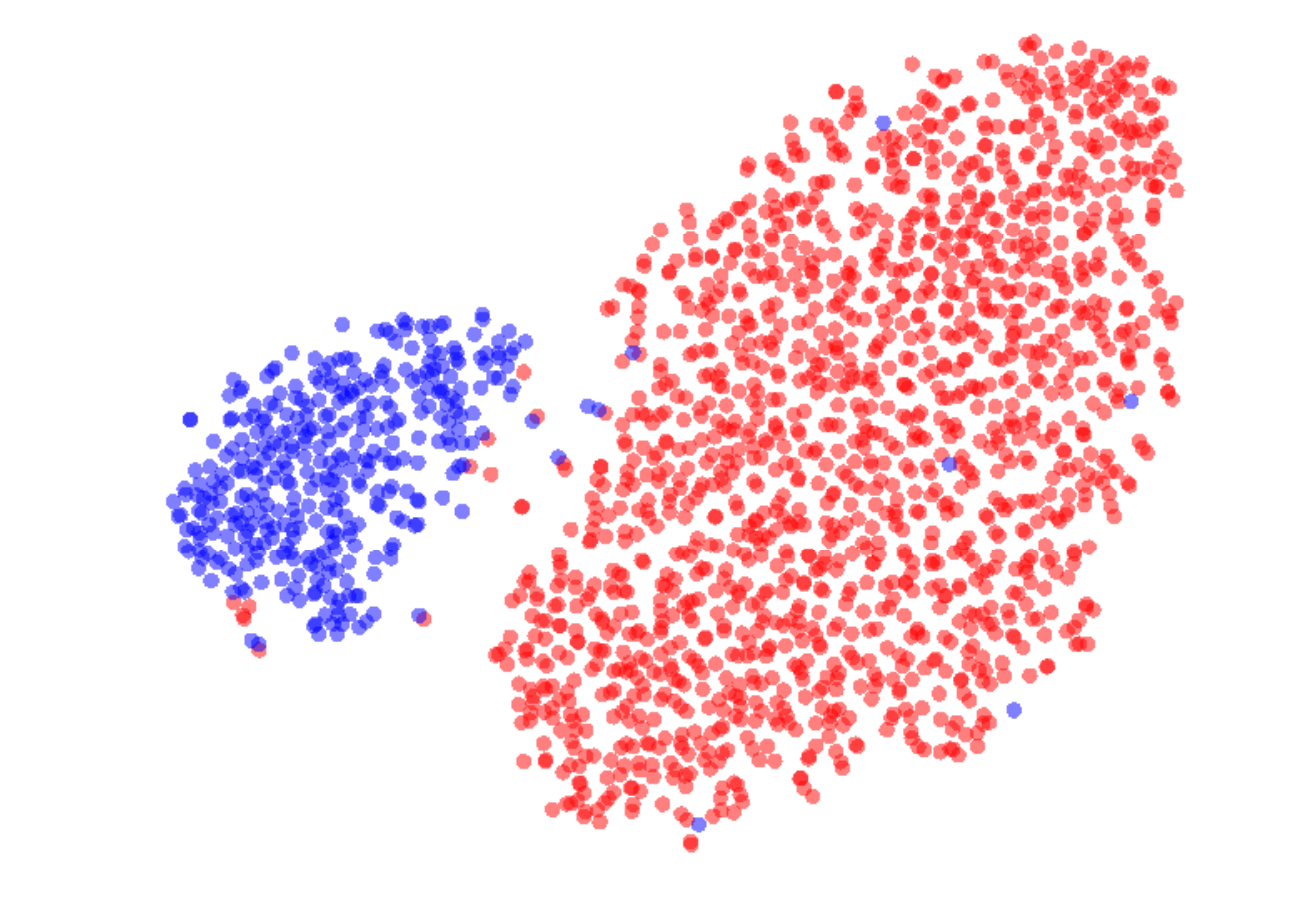}}~~~~~~~~
\subfigure[Deep layer distributions]
{\includegraphics[width=0.35\linewidth]{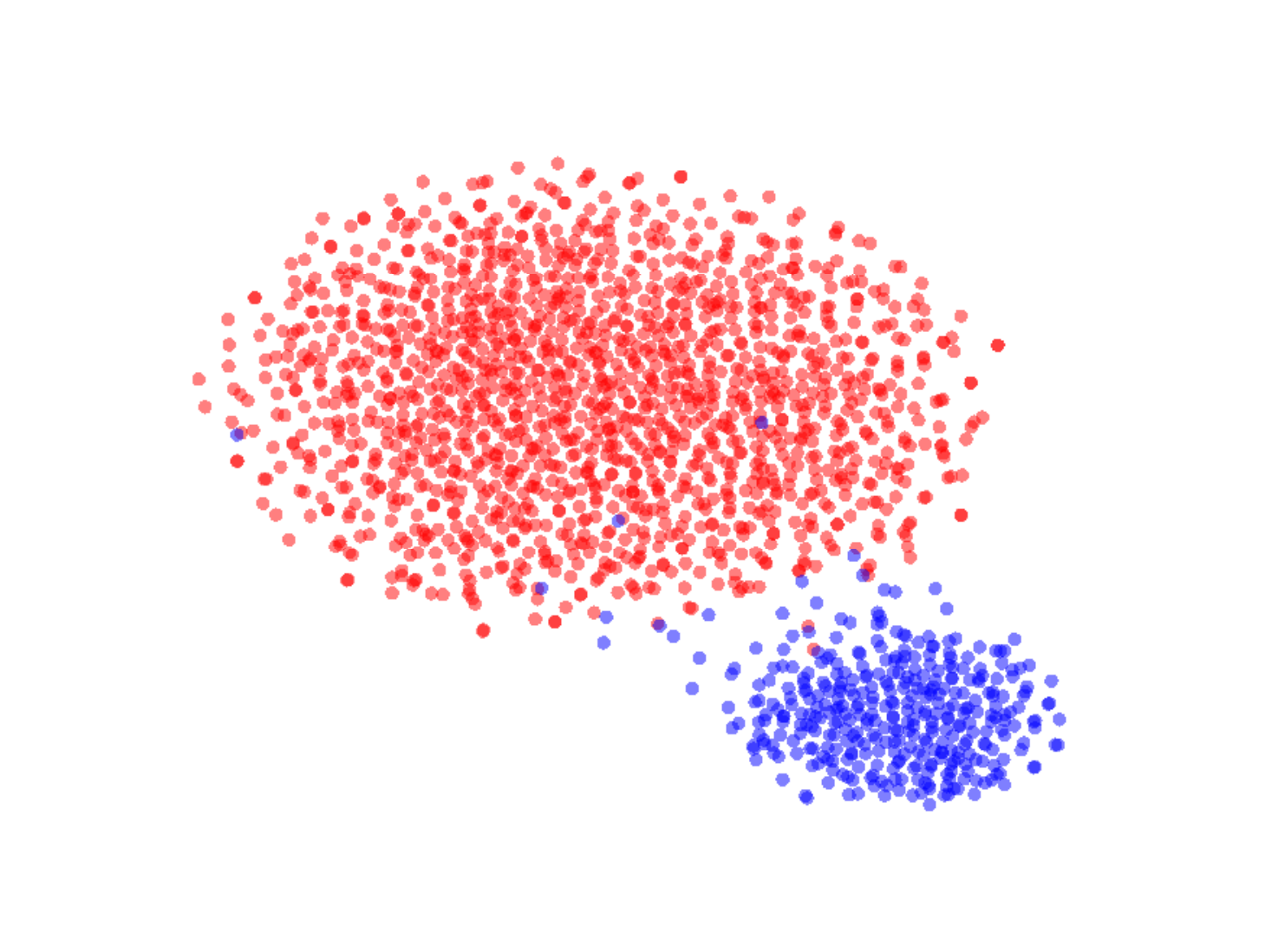}}
\caption{t-SNE~\citep{tsne} visualization of the mini-batch BN feature vector distributions in both shallow and deep layers, across different datasets. Each point represents the BN statistics in one mini-batch. Red dots come from Bing domain, while the blue ones are from Caltech-256 domain. The size of each mini-batch is 64.}\label{fig:bn_visualize}
}
\end{figure}

This pilot experiment suggests:
\begin{enumerate}
	\item Both shallow layers and deep layers of the DNN are influenced by domain shift. Domain adaptation by manipulating the output layer alone is not enough.
	\item The statistics of BN layer contain the traits of the data domain.
	%The statistics of BN layer is a good indication of domains. This confirms the hypothesis that the domain specific knowledge is stored in the statistics of BN layers.
\end{enumerate}

Both observations motivate us to adapt the representation across different domains by BN layer.
\end{subsection}

\begin{subsection}{Adaptive Batch Normalization}\label{sec:method}
Given the pre-trained DNN model and a target domain, our Adaptive Batch Normalization algorithm is as follows\footnote{In practice we adopt an online algorithm~\citep{donald1999art} to efficiently estimate the mean and variance.}:
\floatstyle{plain}
\newfloat{myalgo}{tbhp}{mya}
\begin{myalgo}
\centering
\begin{minipage}{8cm}
\begin{algorithm}[H]
\caption{Adaptive Batch Normalization (AdaBN)}
\begin{algorithmic}
\FOR{neuron $j$ in DNN}
\STATE Concatenate neuron responses on all images of target domain $t$: $\mathbf{x}_j = [\ldots, x_j(m), \ldots]$
\STATE Compute the mean and variance of the target domain: $\mu_j^t = \mathbb{E}(\mathbf{x}_j^t)$, $\sigma_j^t = \sqrt{\text{Var}(\mathbf{x}_j^t)}$.
\ENDFOR

\FOR{neuron $j$ in DNN, testing image $m$ in target domain}
\STATE Compute BN output $y_j(m):= \gamma_j \frac{\big(x_j(m) - \mu_j^t\big)}{\sigma_j^t} + \beta_j$
\ENDFOR
\end{algorithmic}
\end{algorithm}
\end{minipage}
\end{myalgo}

The intuition behind our method is straightforward: The standardization of each layer by domain ensures that each layer receives data from a similar distribution, no matter it comes from the source domain or the target domain.

For $K$ domain adaptation where $K > 2$, we standardize each sample by the statistics in its own domain. During training, the statistics are calculated for every mini-batch, the only thing that we need to make sure is that the samples in every mini-batch are from the same domain. 

For (semi-)supervised domain adaptation, we may use the labeled data to fine-tune the weights as well. As a result, our method could fit in all different settings of domain adaptation with minimal  effort.

%\emph{Then for each BN layer in the model, we replace the statistics (mean and variance) on source domain with those on target domain.} In contrary to other deep learning based domain adaptation methods that modify the structure of network or add new adaptation layers, our method only changes the statistics of BN layer. The intuition behind our method is very straightforward: The standardization of each layer by domain ensures that each layer received data from a similar distribution no matter in source domain or target domain, thus the domain information is anonymized.
%
%For multi-source domain adaptation, the only change is we must standardize each sample by the statistics in its own domain. Since in training the statistics are calculated in one mini-batch, we only need to make sure that the samples in one mini-batch are from the same domain. While for (semi-)supervised domain adaptation, we may use the labeled data to fine-tune the weights as well. As a result, our method could fit in all different settings of domain adaptation with minor effort. 
\end{subsection}

\begin{subsection}{Further Thoughts About Adabn}\label{sec:discuss}
The simplicity of AdaBN is in sharp contrast to the complication of the domain shift problem. One natural question to ask is whether such simple translation and scaling operations  could approximate the intrinsically non-linear domain transfer function. 

Consider a simple neural network with input $\x \in \BR^{p_1 \times 1}$. It has one BN layer with mean and variance of each feature being $\mu_i$ and $\sigma^2_i$ ($i \in \{1 \ldots p_2\}$), one fully connected layer with weight matrix $\W \in \BR^{p_1 \times p_2}$ and bias $\bb \in \BR^{p_2 \times 1}$, and a non-linear transformation layer $f(\cdot)$, where $p_1$ and $p_2$ correspond to the input and output feature size. The output of this network is $f(\W_a \x + \bb_a)$, where
\begin{equation}
	\begin{aligned}
		\W_a &= \W^T \Si^{-1},&  		\bb_a &= -\W^T \Si^{-1} \muu + \bb, \\
		\Si &= \diag(\sigma_1, ..., \sigma_{p_1}),& 	\muu &= (\mu_1, ..., \mu_{p_1}).
	\end{aligned}
\end{equation}
The output without BN is simply $f(\W^T \x + \bb)$. We can see that the transformation is highly non-linear even for a simple network with one computation layer. As CNN architecture goes deeper, it will gain increasing power to represent more complicated transformations.

Another question is why we transform the neuron responses independently, not decorrelate and then re-correlate the responses as suggested in~\citet{coral}. Under certain conditions, decorrelation could improve the performance. However, in CNN, the mini-batch size is usually smaller than the feature dimension, leading to singular covariance matrices that is hard to be inversed. As a result, the covariance matrix is always singular. In addition, decorrelation requires to compute the inverse of the covariance matrix which is computationally intensive, especially if we plan to apply AdaBN to all layers of the network.
\end{subsection}

\end{section}

\begin{section}{Experiments}\label{sec:exp}

In this section, we demonstrate the effectiveness of AdaBN on standard domain adaptation datasets, and empirically analyze the adapted features. We also evaluation our method on a practical application with remote sensing images.   %In the sequel, we refer our method as ``BN Adapt'' for short.

\subsection{Experimental Settings}
We first introduce our experiments on two standard datasets: Office~\citep{office} and Caltech-Bing~\citep{bing-caltech}.

\textbf{Office}~\citep{office} is a standard benchmark for domain adaptation, which is a collection of 4652 images in 31 classes from three different domains: \textit{Amazon}(\textbf{A}), \textit{DSRL}(\textbf{D}) and \textit{Webcam}(\textbf{W}). Similar to~\citep{ddc,coral,dan}, we evaluate the pairwise domain adaption performance of AdaBN on all six pairs of domains. For the multi-source setting, we evaluate our method on three transfer tasks \{\textbf{A, W}\} $\rightarrow$ \textbf{D}, \{\textbf{A, D}\} $\rightarrow$ \textbf{W}, \{\textbf{D, W}\} $\rightarrow$ \textbf{A}.

\textbf{Caltech-Bing}~\citep{bing-caltech} is a much larger domain adaptation dataset, which contains 30,607 and 121,730 images in 256 categories from two domains Caltech-256(\textbf{C}) and Bing(\textbf{B}). The images in the Bing set are collected from  Bing image search engine by keyword search. Apparently Bing data contains noise, and its data distribution is dramatically different from that of Caltech-256.

We compare our approach with a variety of methods, including four shallow methods: SA~\citep{sa}, LSSA~\citep{aljundi2015landmarks}, GFK~\citep{gfk}, CORAL~\citep{coral}, and four deep methods: DDC~\citep{ddc}, DAN~\citep{dan}, RevGrad~\citep{revgrad}, Deep CORAL~\citep{sun2016deep}. Specifically, GFK models domain shift by integrating an infinite number of subspaces that characterize changes in statistical properties from the source to the target domain. SA, LSSA and CORAL align the source and target subspaces by explicit feature space transformations that would map source distribution into the target one. DDC and DAN are deep learning based methods which maximize domain invariance by adding to AlexNet one or several adaptation layers using MMD. RevGrad incorporates a gradient reversal layer in the deep model to encourage learning domain-invariant features. Deep CORAL extends CORAL to perform end-to-end adaptation in DNN. It should be noted that these deep learning methods have the adaptation layers on  top of the output layers of DNNs, which is a sharp contrast to our method that delves into early convolution layers as well with the help of BN layers.

We follow the full protocol~\citep{decaf} for the single source setting; while for multiple sources setting, we use all the samples in the source domains as training data, and use all the samples in the target domain as testing data. We fine-tune the Inception-BN~\citep{bn} model on source domain in each task for 100 epochs. The learning rate is set to $0.01$ initially, and then is dropped by a factor $0.1$ every 40 epochs. Since the office dataset is quite small, following the best practice in~\citet{dan}, we freeze the first three groups of Inception modules, and set the learning rate of fourth and fifth group one tenth of the base learning rate to avoid overfitting. For Caltech-Bing dataset, we fine-tune the whole model with the same base learning rate.

% We use the standard dataset Office~\citep{office} for evaluating domain adaptation algorithms. The dataset consists of 31 classes in three domains: Amazon, DSLR and webcam. We evaluate all the six adaptation settings in our experiments. We fine-tune the CNN on source domain in each setting for 20 epochs. The learning rate starts from $0.01$, and then decreases to $0.001$ after tenth epoch. Since the office dataset is quite small, following the best practice in~\citep{dan}, we freeze the first three groups of Inception modules, and set the learning rate of fourth and fifth group one tenth of the base learning rate. We compare our method with three shallow methods: SA~\citep{sa}, GFK~\citep{gfk}, CORAL~\citep{coral}, and three deep methods: DDC~\citep{ddc}, DAN~\citep{dan}, RevGrad~\citep{revgrad}. In the single source setting, we follow the full protocol in~\citep{decaf}; while for multiple sources setting, we use all the samples in the source domains as training data, and use all the samples in the target domain as testing data.

\subsection{Results}
\subsubsection{Office Dataset}
Our results on Office dataset is reported in Table~\ref{tbl:result} and Table~\ref{tbl:multi} for single/multi source(s), respectively. Note that the first 5 models of the Table~\ref{tbl:result} are pre-trained on AlexNet~\citep{alexnet} instead of the Inception-BN~\citep{bn} model, due to the lack of publicly available pre-trained Inception BN model in Caffe~\citep{caffe}. Thus, the relative improvements over the baseline (AlexNet/Inception BN) make more sense than the absolute numbers of each algorithm.

\begin{table*}[!t]
\small
	\begin{center}
	\setlength{\tabcolsep}{3pt}
	\begin{tabular}{lccccccc}
		\hline	
		Method				&A $\rightarrow$ W		& D $\rightarrow$ W 	& W $\rightarrow$ D 	& A $\rightarrow$ D 	& D $\rightarrow$ A 	& W $\rightarrow$ A 	& Avg	\\
		\hline
		AlexNet~\citep{alexnet}		&61.6	&95.4	&99.0	&63.8	&51.1	&49.8	&70.1\\
		DDC~\citep{ddc} 		 		&61.8	&95.0	&98.5	&64.4	&52.1 	&52.2	&70.6\\
		DAN~\citep{dan} 				&68.5	&96.0	&99.0	&67.0	&54.0	&53.1	&72.9\\
		Deep CORAL~\citep{sun2016deep} &66.4  &95.7  &99.2   &66.8   &52.8   &51.5   &72.1\\
		RevGrad~\citep{revgrad}		&67.3	&94.0	&93.7	&- 		&-		&-		&-	\\
		\hline \hline
%		Inception BN~\citep{bn} 		&66.2	&94.1	&98.4	&70.1	&58.6	&57.5	&74.2\\
		Inception BN~\citep{bn}		&70.3	&94.3	&\textbf{100}		&70.5	&\textbf{60.1}	&57.9	&75.5\\
		SA~\citep{sa} 				&69.8 	&95.5	&99.0	&71.3	&59.4	&56.9	&75.3\\
		GFK~\citep{gfk}	 			&66.7	&\textbf{97.0}	&99.4	&70.1	&58.0	&56.9	&74.7\\
		LSSA~\citep{aljundi2015landmarks} &67.7 &96.1 &98.4  &71.3   &57.8   &57.8   &74.9\\
		CORAL~\citep{coral} 			&70.9	&95.7	&99.8	&71.9	&59.0	&60.2	&76.3\\
		%Adapt BN 						&73.6	&95.8	&99.8	&72.5	&58.6	&56.2	&76.1\\
		%Adapt BN + SA					&73.5	&96.1	&99.2	&73.3	&59.0	&56.0	&76.2\\
		AdaBN					&74.2	&95.7	&99.8	&\textbf{73.1}	&59.8	&57.4	&76.7	\\
		AdaBN + CORAL			&\textbf{75.4}	&96.2	&99.6	&72.7	&59.0	&\textbf{60.5}	&\textbf{77.2}\\
		\hline
	\end{tabular}

	\caption{Single source domain adaptation results on Office-31~\citep{office} dataset with standard unsupervised adaptation protocol.} \label{tbl:result}
	\end{center}
\end{table*}

From Table~\ref{tbl:result}, we first notice that the Inception-BN indeed improves over the AlexNet on average, which means that the CNN pre-trained on ImageNet has learned  general features, the improvements on ImageNet can be transferred to new tasks. Among the methods based on Inception-BN features, our method improves the most over the baseline. Moreover, since our method is complementary to other methods, we can simply apply CORAL on the top of AdaBN. Not surprisingly, this simple combination exhibits 0.5\% increase in performance. This preliminary test reveals further potential of AdaBN if combined with other advanced domain adaptation methods. Finally, we could improve 1.7\% over the baseline, and advance the state-of-the-art results for this dataset. %Compared to other methods based on AlexNet, our method is better than DDC and RevGrad, and worse than DAN and Deep CORAL in terms of relative improvements over corresponding baselines.
%In terms of relative improvements over corresponding baselines, DAN has larger gain than other methods.

None of the compared methods has reported their  performance on multi-source domain adaptation. To demonstrate the capacity of AdaBN under multi-domain settings, we compare it against CORAL, which is the best performing algorithm in the single source setting. % here we only compare AdaBN with the best algorithm CORAL in the single source setting. Analyzing the results of the baseline in Table~\ref{tbl:multi}, 
The result is reported in Table~\ref{tbl:multi}. We find that simply combining two domains does not lead to better performance. The result is generally worse compared to the best performing single domain between the two. This phenomenon suggests that if we cannot properly cope with domain bias, the increase of training samples may be reversely affect to the testing performance. This result confirms the necessity of domain adaptation. In this more challenging setting, AdaBN still outperforms the baseline and CORAL on average. Again, when combined with CORAL, our method demonstrates further improvements. At last, our method archives 2.3\% gain over the baseline.

\begin{table*}[!htb]
\small
	\begin{center}
	\begin{tabular}{lcccc}
		\hline	
		Method				&A, D $\rightarrow$ W		&A, W $\rightarrow$ D 	& D, W $\rightarrow$ A 	& Avg	\\
		\hline
		Inception BN~\citep{bn} 		&90.8	&95.4	&60.2	&82.1\\
		CORAL~\citep{coral} 		&92.1		&96.4	&\textbf{61.4}	&83.3\\
		AdaBN 					&94.2	&97.2	&59.3	&83.6\\
		AdaBN + CORAL			&\textbf{95.0}	&\textbf{97.8}	&60.5	&\textbf{84.4}\\
		\hline
	\end{tabular}
	
	\caption{Multi-source domain adaptation results on Office-31~\citep{office} dataset with standard unsupervised adaptation protocol.} \label{tbl:multi}

	\end{center}
\end{table*}

\subsubsection{Caltech-Bing Dataset}
To further evaluate our method on the large-scale dataset, we show our results on Caltech-Bing Dataset in Table~\ref{tbl:caltech-bing}. Compared with CORAL, AdaBN achieves better performance, which improves 1.8\% over the baseline. Note that all the domain adaptation methods show minor improvements over the baseline in the task \textbf{C} $\rightarrow$ \textbf{B}. One of the hypotheses to this relatively small improvement is that the images in Bing dataset are collected from Internet, which are more diverse and noisier~\citep{bing-caltech}. Thus, it is not easy to adapt on the Bing dataset from the relatively clean dataset Caltech-256. Combining CORAL with our method does not offer further improvements. This might be explained by  the noise of the Bing dataset and the imbalance of the number of images in the two domains.

\begin{table}[!htb]
\small
	\begin{center}
	\setlength{\tabcolsep}{3pt}
	\begin{tabular}{lccc}
		\hline	
		Method					&C $\rightarrow$ B		& B $\rightarrow$ C 	& Avg	\\
		\hline
		Inception BN~\citep{bn} 	&35.1	&64.6		&49.9\\
		CORAL~\citep{coral} 		&\textbf{35.3}		&67.2		&51.3\\
		AdaBN 					&35.2	&\textbf{68.1}		&\textbf{51.7}\\
		AdaBN + CORAL				&35.0		&67.5		&51.2\\
		\hline
	\end{tabular}

	\caption{Single source domain adaptation results on Caltech-Bing~\citep{bing-caltech} dataset.} \label{tbl:caltech-bing}

	\end{center}
\end{table}

\subsection{Empirical Analysis}
In this section, we empirically analyze the features adapted by our method and investigate the influence of the number of samples in target domain to the performance.

% \subsubsection{Feature Visualization.}
% We first visualize the features of the last layer before and after adaptation using t-SNE~\citep{tsne} in Fig.~\ref{fig:visualization}. We choose two adaption settings for illustration: Amazon to Webcam and Amazon to DSLR. Each red circle represents one training sample, while each blue circle represents one testing sample. We can see that the features of testing data after adaption are blended more evenly with the training data compared to those without adaption. In other words, the distribution of testing samples is more consistent with the training one. This intuitive illustration again confirms that our method is effective against domain shift.
% \begin{figure*}[htb]
% \begin{center}
% 	\subfigure[A $\rightarrow$ W, with adaptation]{\includegraphics[width=0.24\linewidth]{a2w_adapt}}~
% 	\subfigure[A $\rightarrow$ W, w/o adaptation]{\includegraphics[width=0.24\linewidth]{a2w_no_adapt}}~ 
% 	\subfigure[A $\rightarrow$ D, with adaptation]{\includegraphics[width=0.24\linewidth]{a2d_adapt}}~
% 	\subfigure[A $\rightarrow$ D, w/o adaptation]{\includegraphics[width=0.24\linewidth]{a2d_no_adapt}}
% 	\caption{Visualization of features in the last layer. Red circles are training samples, while blues ones are testing samples. Best viewed in color.} \label{fig:visualization}
% \end{center}
% \end{figure*}

\subsubsection{Analysis of Feature Divergence.}
In this experiment, we analyze the statistics of the output of one shallow layer (the output of second convolution layer) and one deep layer (the output of last Inception module before ReLU) in the network. In particular, we compute the distance of source domain distribution and target domain distribution before and after adaptation. We denote each feature $i$ as $F_i$, and assume that the output of each feature generally follows a Gaussian distribution with mean $\mu_i$ and variance $\sigma_i^2$. Then we use the symmetric KL divergence as our metric:
\begin{equation}
	\begin{aligned}
		D(F_i \mid\mid F_j) &= \text{KL}(F_i \mid \mid F_j) + \text{KL}(F_j \mid \mid F_i),\\
		\text{KL}(F_i \mid \mid F_j) &= \log \frac{\sigma_j}{\sigma_i} + \frac{\sigma_i^2 + (\mu_i - \mu_j)^2}{2\sigma_j^2} - \frac{1}{2}.
	\end{aligned}
\end{equation}
We plot the distribution of the distances in Fig.~\ref{fig:distribution}. Our method reduces the domain discrepancy in both shallow layer and deep layer. We also report the quantitative results in Table.~\ref{tbl:distribution}. This experiment once again verifies the effectiveness of the proposed method.
\begin{figure*}[!htb]
\begin{center}
	\subfigure[A $\rightarrow$ W, shallow layer]{\includegraphics[width=0.24\linewidth]{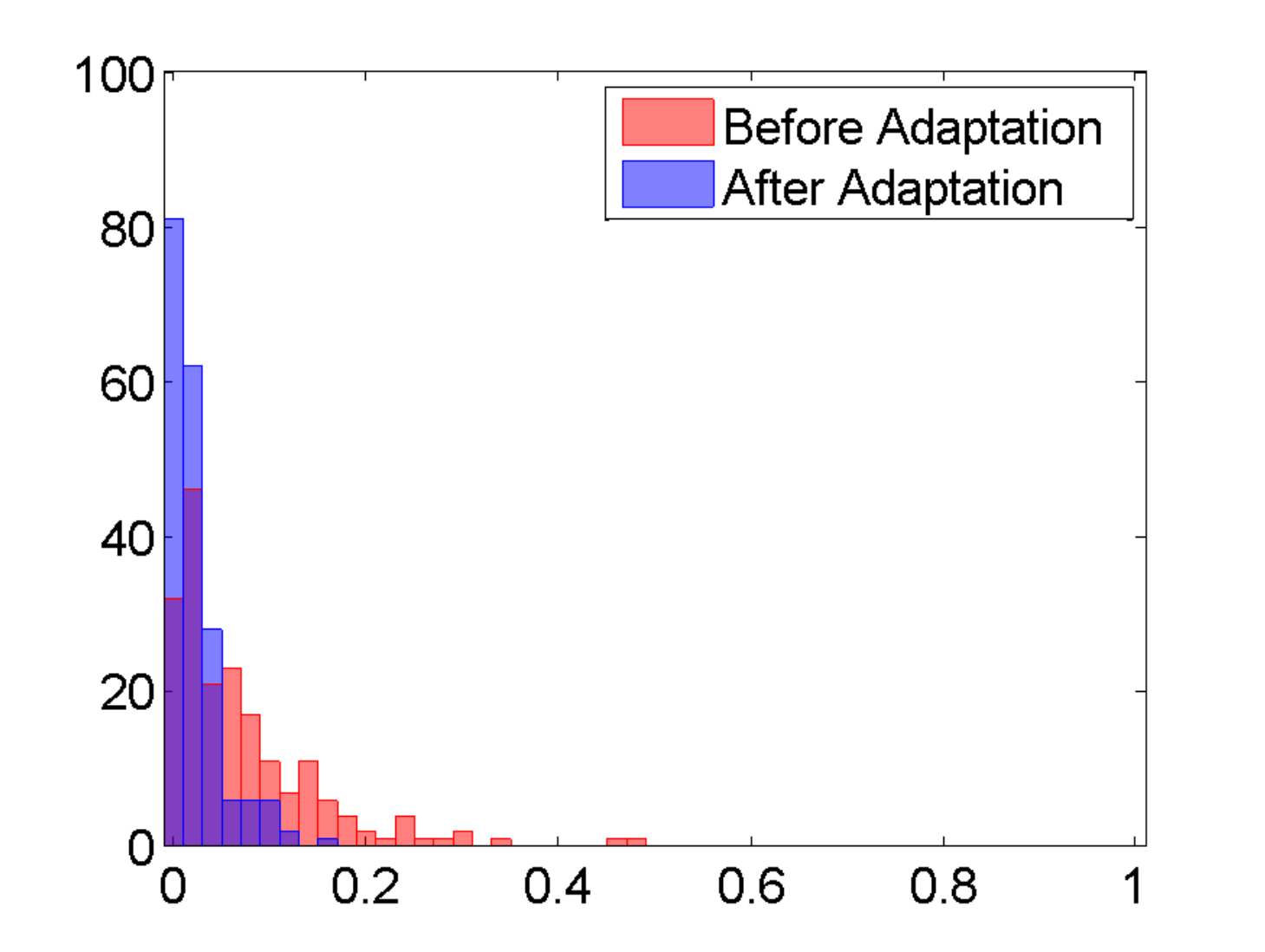}}~
	\subfigure[A $\rightarrow$ W, deep layer]{\includegraphics[width=0.24\linewidth]{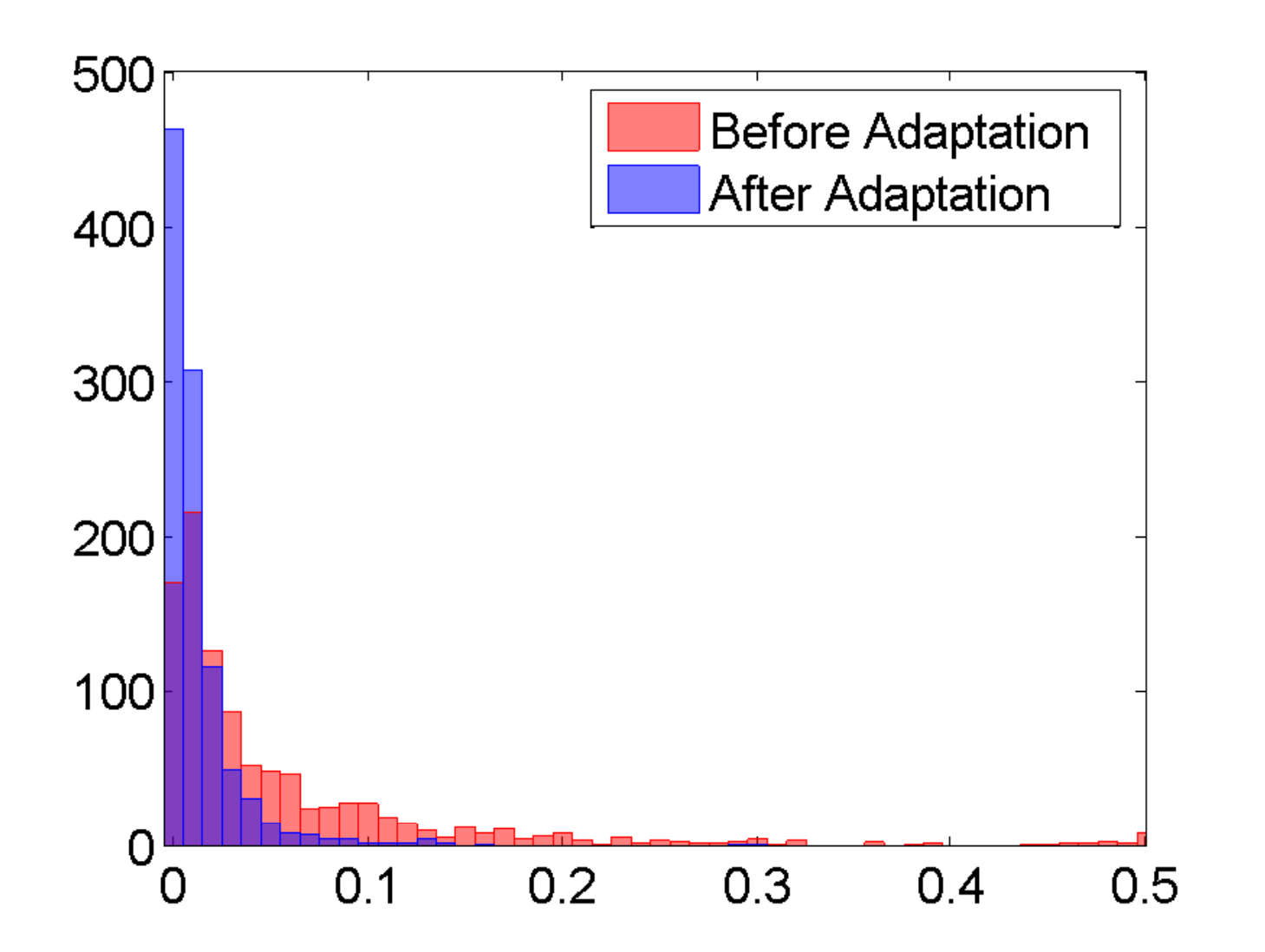}}~
	\subfigure[A $\rightarrow$ D, shallow layer]{\includegraphics[width=0.24\linewidth]{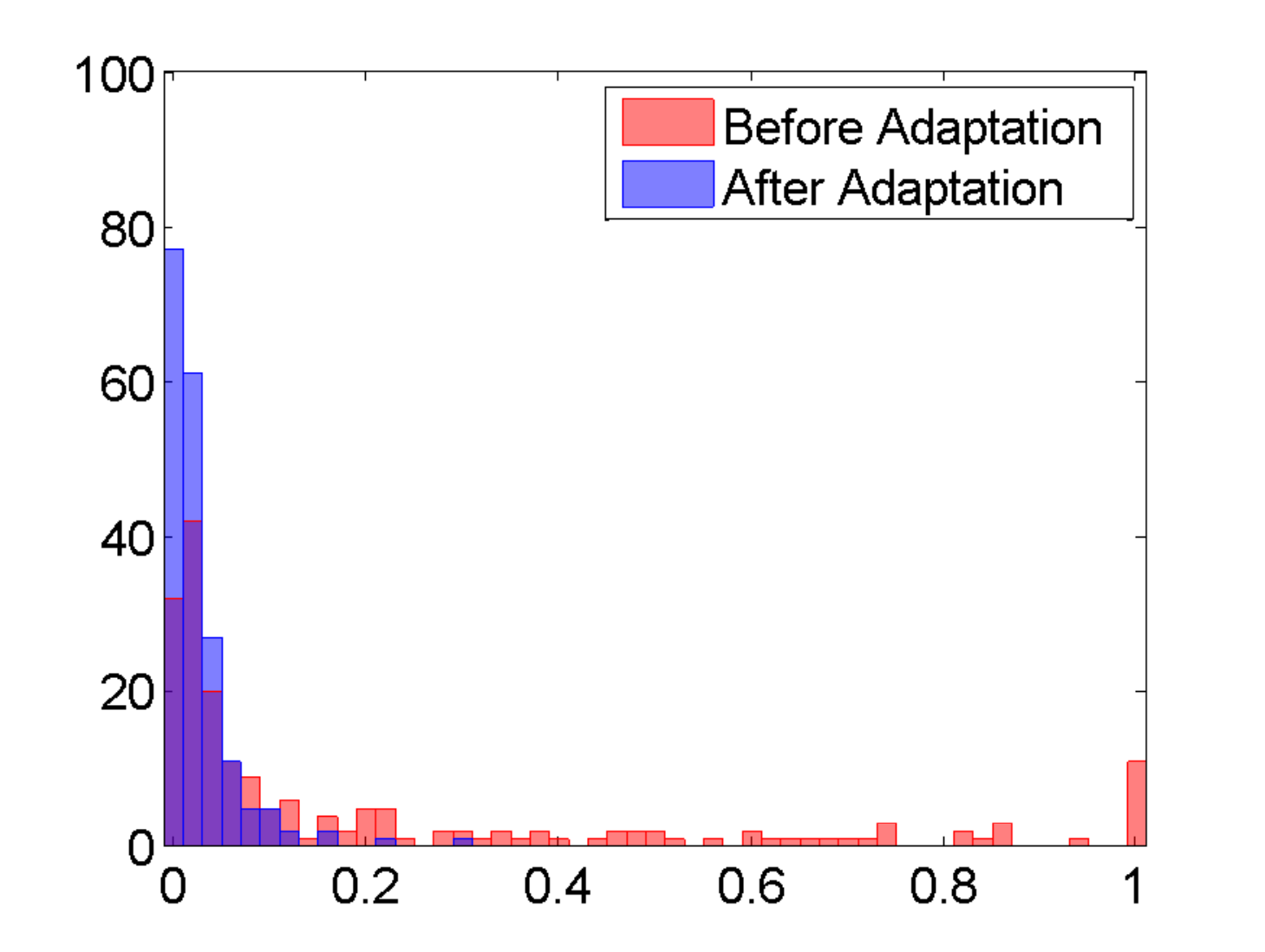}}~
	\subfigure[A $\rightarrow$ D, deep layer]{\includegraphics[width=0.24\linewidth]{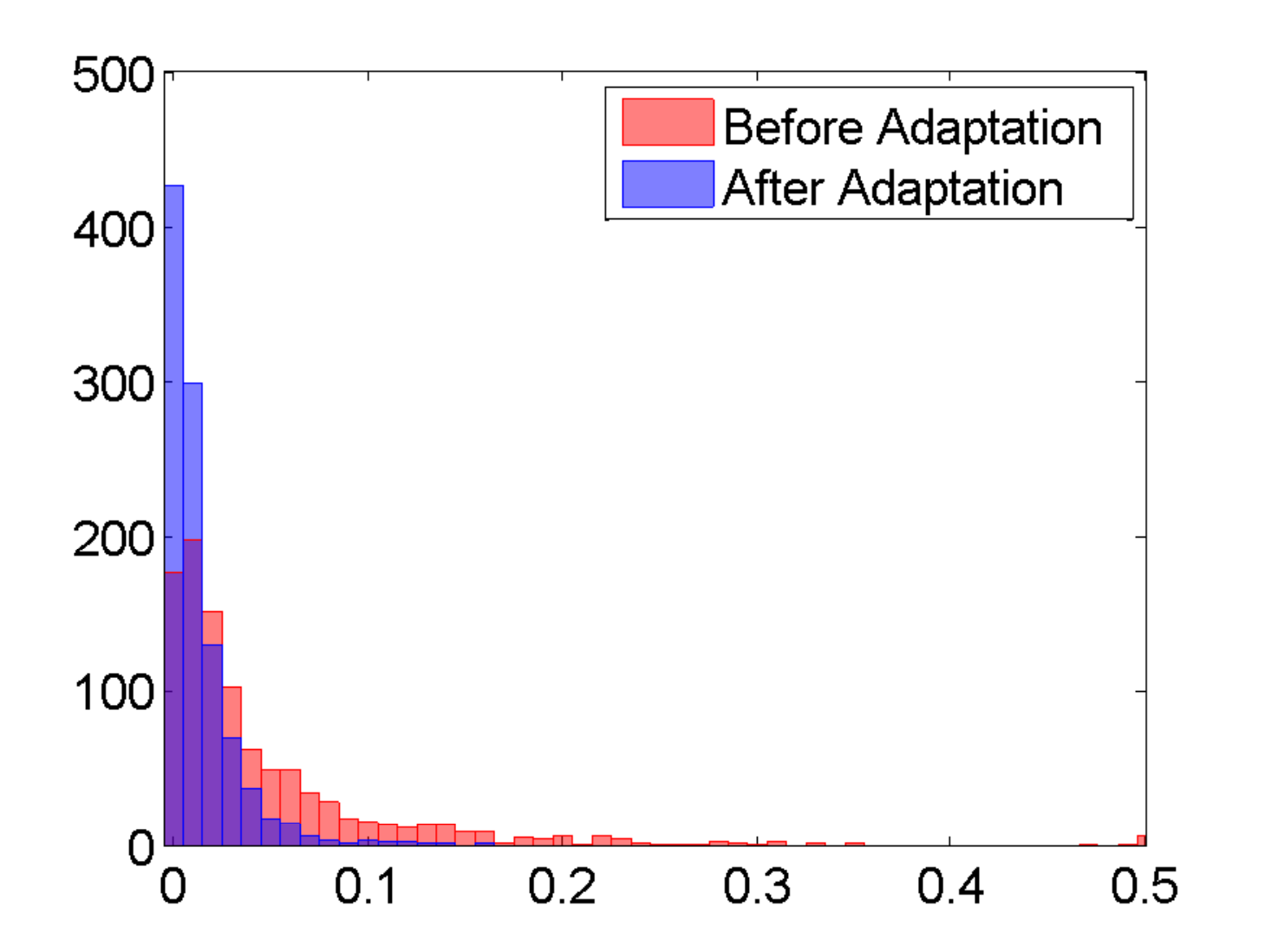}}
	\caption {Distribution of the symmetric KL divergence of the outputs in shallow layer and deep layer. Best viewed in color.} \label{fig:distribution}
\end{center}
\end{figure*}

%\rowcolors{2}{white}{white}{gray!25}
\begin{table}[!htb]
\small
	\begin{center}
	\setlength{\tabcolsep}{4pt}
	\begin{tabular}{lcccc}

		\hline	
		%				&A $\rightarrow$ W, shallow		&A $\rightarrow$ W, deep 	& A $\rightarrow$ D, shallow 	& A $\rightarrow$ D, deep	\\
		\multirow{2}{*}{}				&A $\rightarrow$ W	&A $\rightarrow$ W	& A $\rightarrow$ D	& A $\rightarrow$ D	\\
		\rowcolor{white} 
		& shallow & deep & shallow & deep \\
		\hline
		\rowcolor{gray!25} 
		Before Adapt			&0.0716	&0.0614	&0.2307	&0.0502\\
		\rowcolor{white}
		After Adapt			&0.0227	&0.0134	&0.0266	&0.0140\\
		\hline
	\end{tabular}

	\caption{The average symmetric KL divergence of the outputs in shallow layer and deep layer, respectively.} \label{tbl:distribution}

	\end{center}
\end{table}

\subsubsection{Sensitivity to Target Domain Size.} Since the key of our method is to calculate the mean and variance of the target domain on different BN layers, it is very natural to ask how many target images is necessary to obtain  stable statistics. In this experiment, we randomly select a subset of images in target domain to calculate the statistics and then evaluate the performance on the whole target set.  Fig.~\ref{fig:target-number} illustrates the effect of using different number of batches. The results demonstrate that our method can obtain good results when using only a small part of the target examples. It should also be noted that in the extremal case of  one batch of target images, our method still achieves better results than the baseline. This is valuable in practical use since a large number of target images are often not available.

\begin{figure}[!htb]
\center{
	\subfigure[A $\rightarrow$ W]{\includegraphics[width=0.38\linewidth]{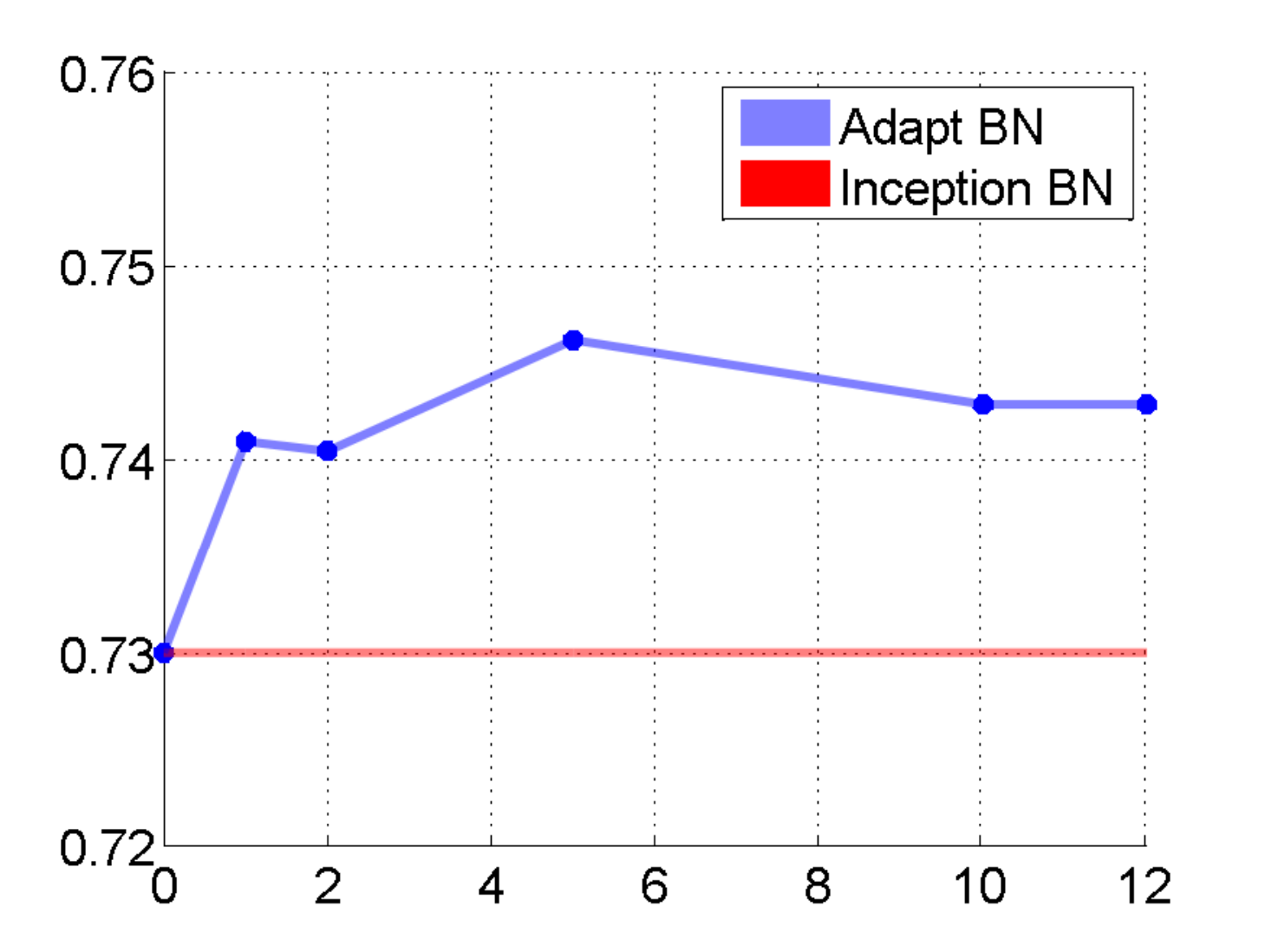}}~~~~
	\subfigure[B $\rightarrow$ C]{\includegraphics[width=0.38\linewidth]{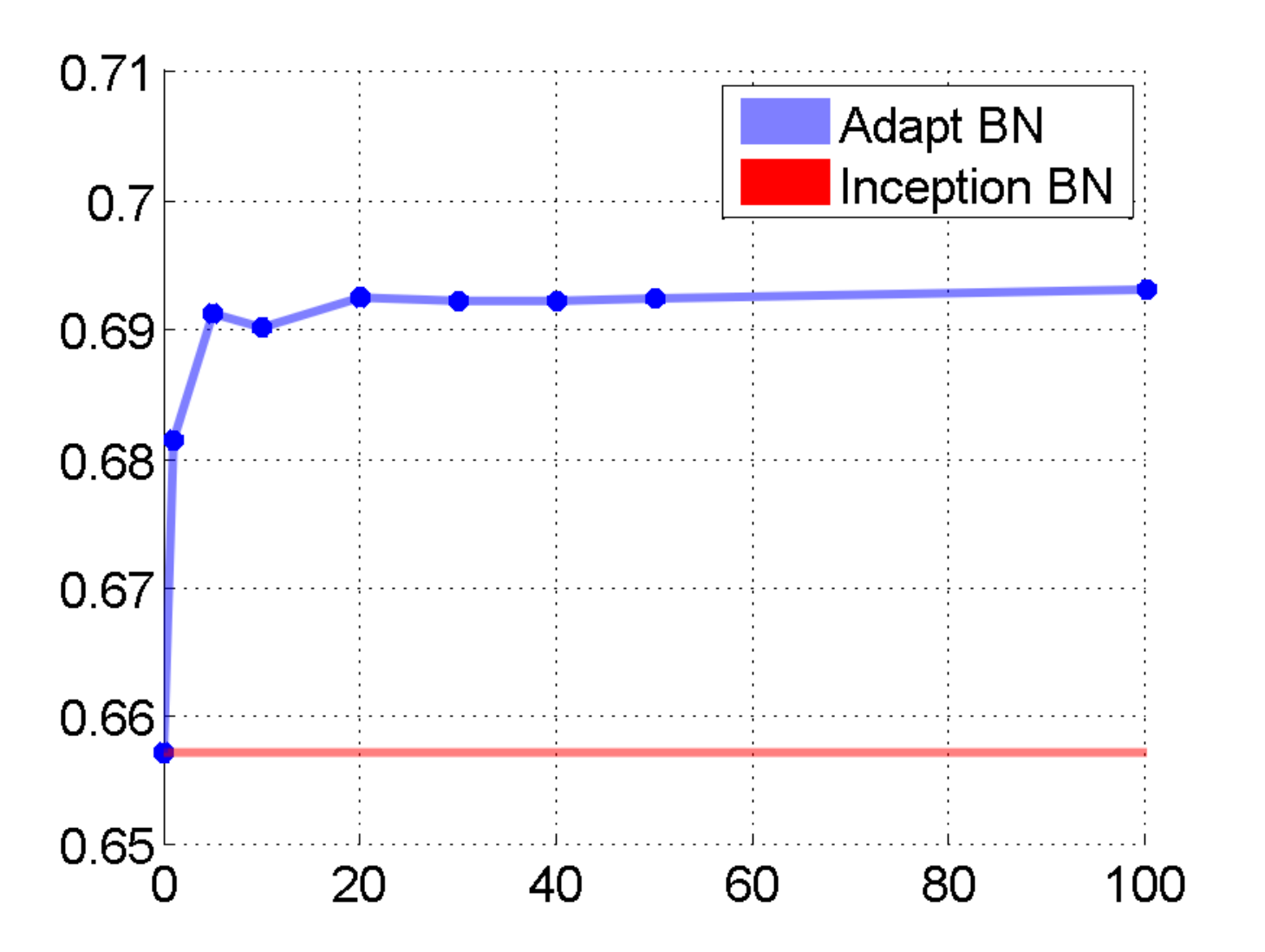}}
	\caption{Accuracy when varying the number of mini-batches used for calculating the statistics of BN layers in \textbf{A} $\rightarrow$ \textbf{W} and \textbf{B} $\rightarrow$ \textbf{C}, respectively. For \textbf{B} $\rightarrow$ \textbf{C}, we only show the results of using less than 100 batches, since the results are very stable when adding more examples. The batch size is 64 in this experiment.
	}\label{fig:target-number}
}
\end{figure}

\begin{subsection}{Practical Application for Cloud Detection in Remote Sensing Images}
In this section, we further demonstrate the effectiveness of AdaBN on a practical problem: Cloud Detection in Remote Sensing Images. Since remote sensing images are taken by different satellites with different sensors and resolutions, the captured images are visually different in texture, color, and value range distributions, as shown in Fig.~\ref{fig:rsimage}. How to adapt a model trained on one satellite to another satellite images is naturally a domain adaptation problem.

Our task here is to identify cloud from the remote sensing images, which can be regarded as a semantic segmentation task. The experiment is taken under a self-collected dataset, which includes three image sets, from GF2, GF1 and Tianhui satellites. Each image set contains 635, 324 and 113 images  with resolution over 6000x6000 pixels respectively. We name the three different datasets following the satellite names. GF2 dataset is used as the training dataset while GF1 and Tianhui datasets are for testing. We use a state-of-art semantic segmentation method \citep{chen2016deeplab} as our baseline model.%Only the GF2 dataset contains labeled training data. We use the training data from GF2 dataset to train a cloud detector for GF2 images. The testing result in GF2 test set can reach to 87.64\% mIOU for our baseline model.

\begin{table}[!htb]
	\begin{center}
	\setlength{\tabcolsep}{3pt}
	\begin{tabular}{lcc}
		\hline	
		Method					& GF1		& Tianhui 	\\
		\hline
		Baseline 	& 38.95\%	& 14.54\%\\
		AdaBN 		& \textbf{64.50\%}	& \textbf{29.66\%}		\\
		\hline
	\end{tabular}

	\caption{Domain adaptation results (mIOU) on GF1 and Tianhui datasets training on GF2 datasets.} \label{tbl:remote}

	\end{center}
\end{table}

The results on GF1 and Tianhui datasets are shown in Table~\ref{tbl:remote}. The relatively low results of the baseline method indicate that there exists large distribution disparity among images from different satellites. Thus, the significant improvement after applying AdaBN reveals the effectiveness of our method. Some of the visual results are shown in Fig.~\ref{fig:rsadabn}. Since other domain adaptation methods require either additional optimization steps and extra components ($e.g.$ MMD) or post-processing distribution alignment (like CORAL), it is very hard to apply these methods from image classification to this large-size (6000x6000) segmentation problem. Comparatively, besides the effective performance, our method needs no extra parameters and very few computations over the whole adaptation process.
% Directly adapting the model trained on GF2 dataset to GF1 test set. The resulted mIOU is only 38.95\%. When applying adaBN to the GF1 test set with 1 epoch to accumulating the statistics on BN, the result improves to 59.03\%. When we update the BN statistics by the test set with 10 epochs, the result further improves to 64.5\%. The significant improvement reveals the effectiveness of our adaBN method. Some of the visual result is shown in Fig.~\ref{fig:rsadabn}

% We further test the adaBN on our next Tianhui dataset. Directly using the model trained on GF2 only reach 14.54\% mIOU. This is because images in Tianhui satellite varies more from GF2 images and the domain variance is larger. After applying adaBN on Tianhui test set with 1 epoch for updating BN statistics, the results imrpoves to 29.07\%. It further improves to 29.66\% with 8 epochs updating. Although with larger domain gap, using adaBN will still improves the final result consistently. 

\begin{figure*}[!htpb]
\begin{center}
	\subfigure[GF1 image]{\includegraphics[width=0.26\linewidth]{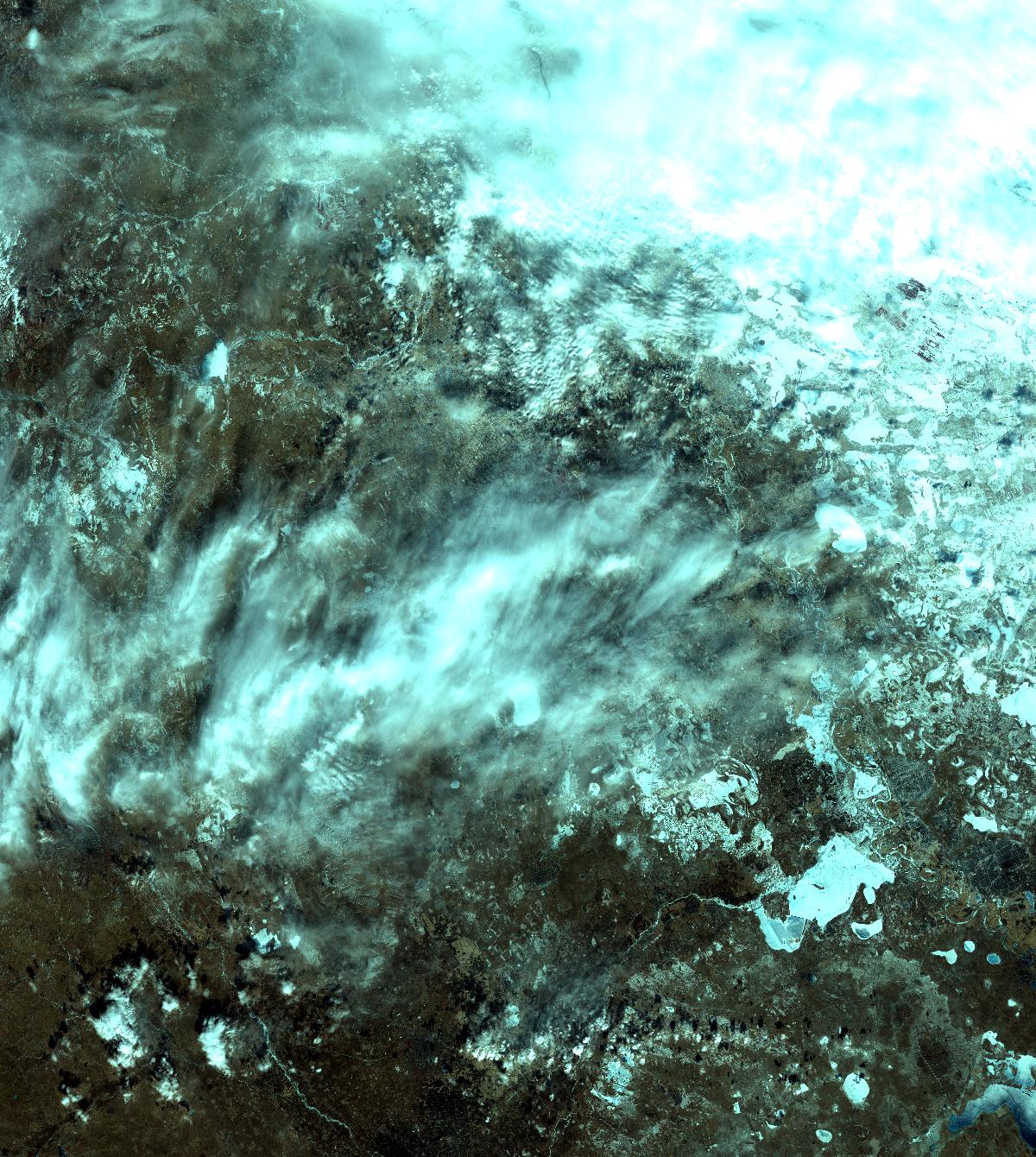}}
	\subfigure[GF2 image]{\includegraphics[width=0.29\linewidth]{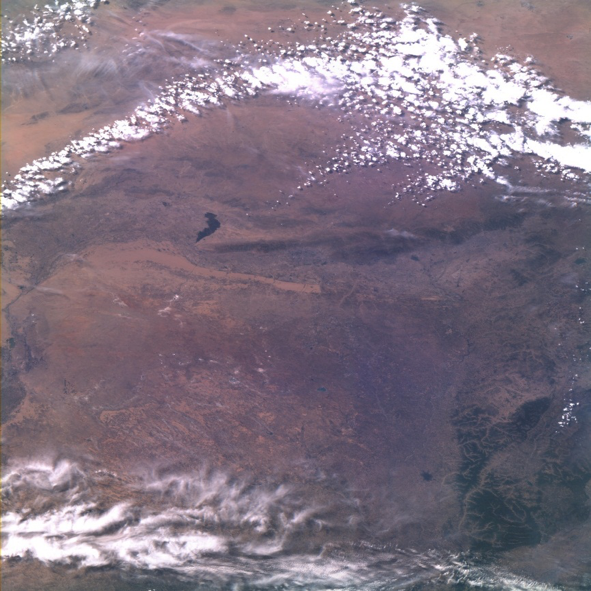}}
	\subfigure[Tianhui image]{\includegraphics[width=0.29\linewidth]{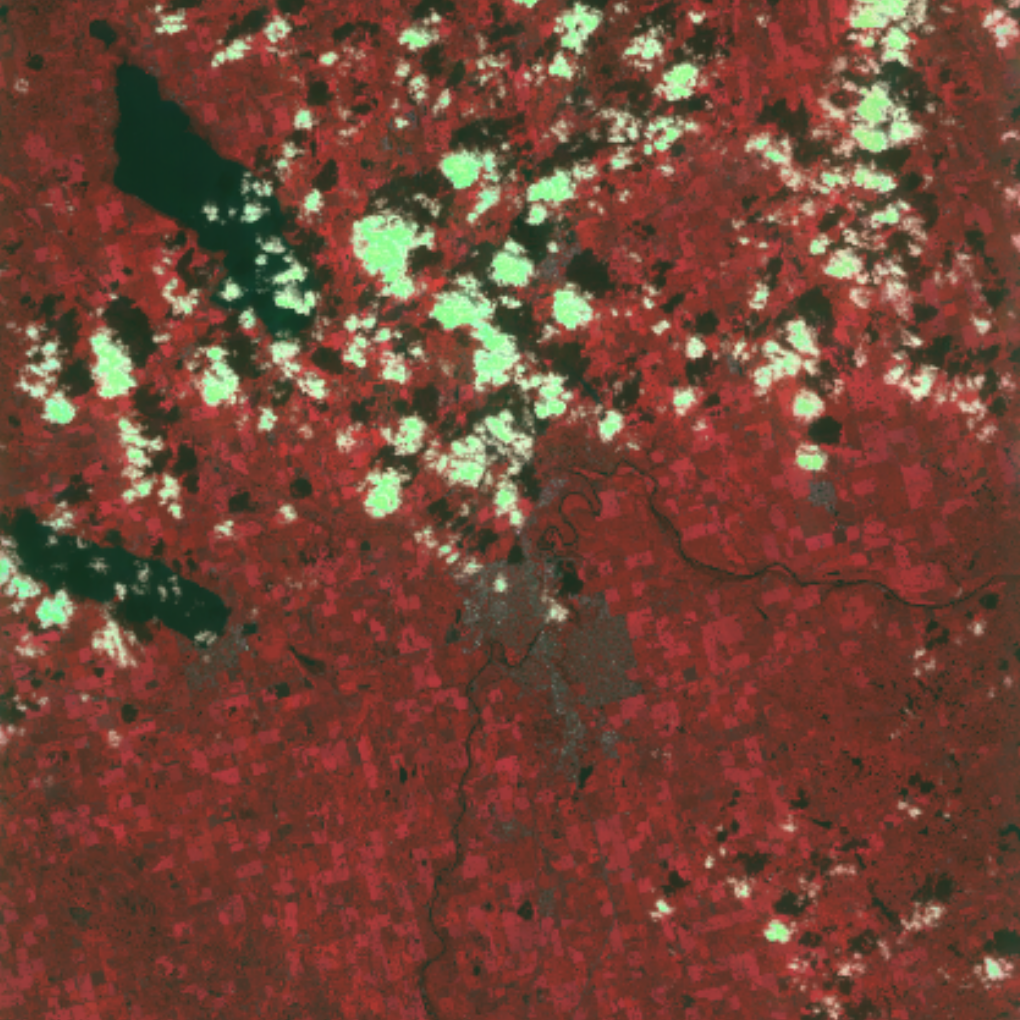}}
	\caption{Remote sensing images in different domains.} \label{fig:rsimage}
\end{center}
\end{figure*}

\begin{figure*}[!htpb]
\begin{center}
	\subfigure[Original image]{\includegraphics[width=0.3\linewidth]{fig/origin_gf2}}
	\subfigure[Without AdaBN]{\includegraphics[width=0.3\linewidth]{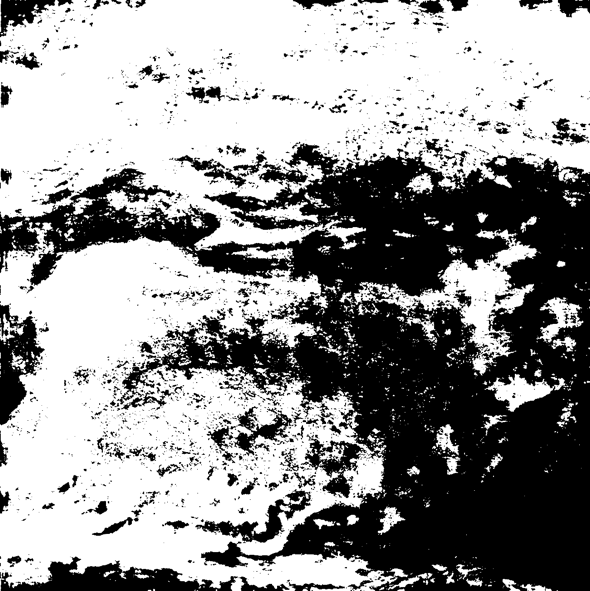}}
	\subfigure[AdaBN]{\includegraphics[width=0.3\linewidth]{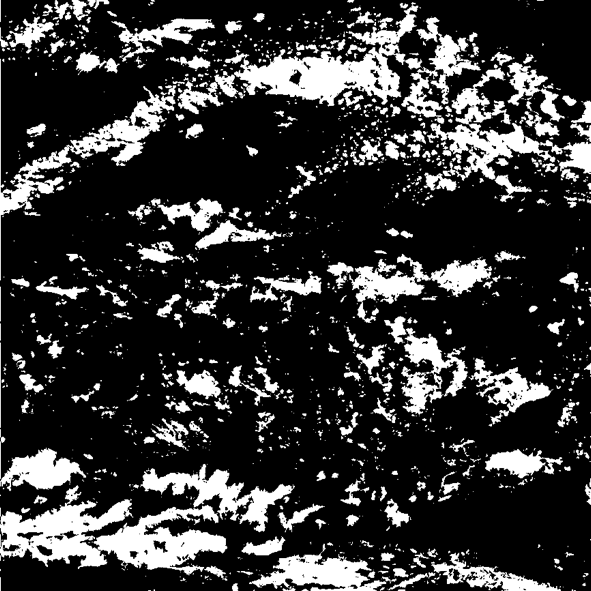}}\\
	\setcounter{subfigure}{0}% Reset subfigure counter
	\subfigure[Original image]{\includegraphics[width=0.3\linewidth]{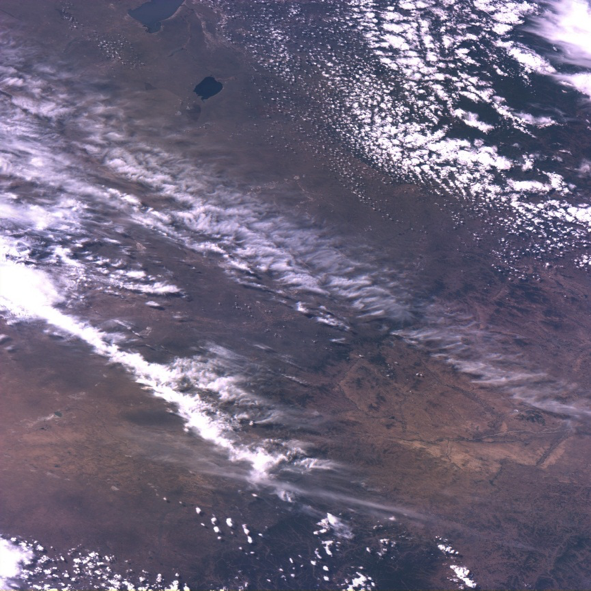}}
	\subfigure[Without AdaBN]{\includegraphics[width=0.3\linewidth]{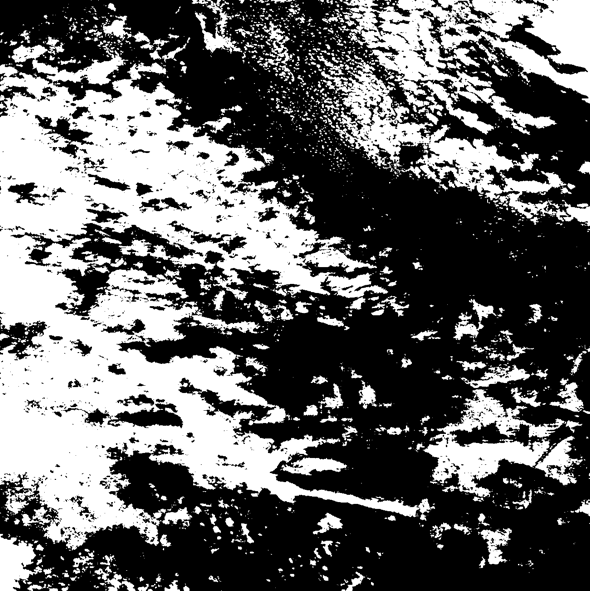}}
	\subfigure[AdaBN]{\includegraphics[width=0.3\linewidth]{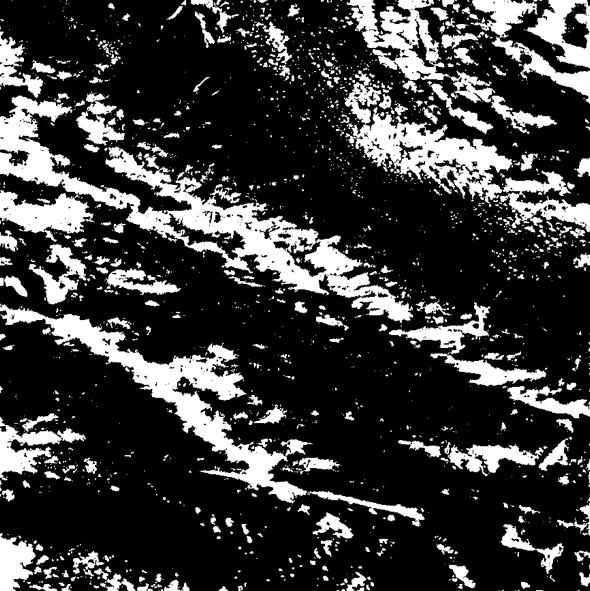}}
\end{center}
	\caption{Visual cloud detection results on GF1 dataset. White pixels in (b) and (c) represent the detected cloud regions.} \label{fig:rsadabn}
\end{figure*}

\end{subsection}

\end{section}

\section{Conclusion and Future Works}
In this paper, we have introduced a simple yet effective approach for domain adaptation on batch normalized neural networks. Besides its original uses, we have exploited another functionality of Batch Normalization (BN) layer: domain adaptation. The main idea is to replace the statistics of each BN layer in source domain with those in target domain. The proposed method is easy to implement and parameter-free, and it takes almost no effort to extend to multiple source domains and semi-supervised settings. % Moreover, our method is not sensitive to the target domain size. Thus it is more favorable for practitioners compared with other deep learning based methods.
Our method established new state-of-the-art results on both single and multiple source(s) domain adaptation settings on standard benchmarks. At last, the experiments on cloud detection for large-size remote sensing images further demonstrate the effectiveness of our method in practical use. We believe our method opens up a new direction for domain adaptation.

In contrary to other methods that use Maximum Mean Discrepancy (MMD) or domain confusion loss to update the weights in CNN for domain adaptation, our method only modifies the statistics of BN layer. Therefore, our method is fully complementary to other existing deep learning based methods. It is interesting to see how these different methods can be unified under one framework.

{%\small
\bibliography{bn}

\begin{thebibliography}{40}
\providecommand{\natexlab}[1]{#1}
\providecommand{\url}[1]{\texttt{#1}}
\expandafter\ifx\csname urlstyle\endcsname\relax
  \providecommand{\doi}[1]{doi: #1}\else
  \providecommand{\doi}{doi: \begingroup \urlstyle{rm}\Url}\fi

\bibitem[Aljundi et~al.(2015)Aljundi, Emonet, Muselet, and
  Sebban]{aljundi2015landmarks}
Rahaf Aljundi, R{\'e}mi Emonet, Damien Muselet, and Marc Sebban.
\newblock Landmarks-based kernelized subspace alignment for unsupervised domain
  adaptation.
\newblock In \emph{CVPR}, 2015.

\bibitem[Baktashmotlagh et~al.(2013)Baktashmotlagh, Harandi, Lovell, and
  Salzmann]{dip}
Mahsa Baktashmotlagh, Mehrtash Harandi, Brian Lovell, and Mathieu Salzmann.
\newblock Unsupervised domain adaptation by domain invariant projection.
\newblock In \emph{ICCV}, pp.\  769--776, 2013.

\bibitem[Beijbom(2012)]{beijbom2012domain}
Oscar Beijbom.
\newblock Domain adaptations for computer vision applications.
\newblock \emph{arXiv preprint arXiv:1211.4860}, 2012.

\bibitem[Bergamo \& Torresani(2010)Bergamo and Torresani]{bing-caltech}
Alessandro Bergamo and Lorenzo Torresani.
\newblock Exploiting weakly-labeled web images to improve object
  classification: a domain adaptation approach.
\newblock In \emph{NIPS}, pp.\  181--189, 2010.

\bibitem[Bousmalis et~al.(2016)Bousmalis, Trigeorgis, Silberman, Krishnan, and
  Erhan]{bousmalis2016domain}
Konstantinos Bousmalis, George Trigeorgis, Nathan Silberman, Dilip Krishnan,
  and Dumitru Erhan.
\newblock Domain separation networks.
\newblock \emph{NIPS}, 2016.

\bibitem[Chen et~al.(2016{\natexlab{a}})Chen, Papandreou, Kokkinos, Murphy, and
  Yuille]{chen2016deeplab}
Liang-Chieh Chen, George Papandreou, Iasonas Kokkinos, Kevin Murphy, and Alan~L
  Yuille.
\newblock Deeplab: Semantic image segmentation with deep convolutional nets,
  atrous convolution, and fully connected crfs.
\newblock \emph{arXiv preprint arXiv:1606.00915}, 2016{\natexlab{a}}.

\bibitem[Chen et~al.(2016{\natexlab{b}})Chen, Li, Li, Lin, Wang, Wang, Xiao,
  Xu, Zhang, and Zhang]{mxnet}
Tianqi Chen, Mu~Li, Yutian Li, Min Lin, Naiyan Wang, Minjie Wang, Tianjun Xiao,
  Bing Xu, Chiyuan Zhang, and Zheng Zhang.
\newblock {MXNet}: A flexible and efficient machine learning library for
  heterogeneous distributed systems.
\newblock \emph{NIPS Workshop on Machine Learning Systems}, 2016{\natexlab{b}}.

\bibitem[Chopra et~al.(2013)Chopra, Balakrishnan, and Gopalan]{dlid}
Sumit Chopra, Suhrid Balakrishnan, and Raghuraman Gopalan.
\newblock {DLID}: Deep learning for domain adaptation by interpolating between
  domains.
\newblock In \emph{ICML Workshop on Challenges in Representation Learning},
  volume~2, 2013.

\bibitem[Donahue et~al.(2014)Donahue, Jia, Vinyals, Hoffman, Zhang, Tzeng, and
  Darrell]{decaf}
Jeff Donahue, Yangqing Jia, Oriol Vinyals, Judy Hoffman, Ning Zhang, Eric
  Tzeng, and Trevor Darrell.
\newblock {DeCAF}: A deep convolutional activation feature for generic visual
  recognition.
\newblock In \emph{ICML}, pp.\  647--655, 2014.

\bibitem[Donald(1999)]{donald1999art}
E~Knuth Donald.
\newblock The art of computer programming.
\newblock \emph{Sorting and searching}, 3:\penalty0 426--458, 1999.

\bibitem[Fernando et~al.(2013)Fernando, Habrard, Sebban, and Tuytelaars]{sa}
Basura Fernando, Amaury Habrard, Marc Sebban, and Tinne Tuytelaars.
\newblock Unsupervised visual domain adaptation using subspace alignment.
\newblock In \emph{ICCV}, pp.\  2960--2967, 2013.

\bibitem[Ganin \& Lempitsky(2015)Ganin and Lempitsky]{revgrad}
Yaroslav Ganin and Victor Lempitsky.
\newblock Unsupervised domain adaptation by backpropagation.
\newblock In \emph{ICML}, pp.\  1180--1189, 2015.

\bibitem[Ghifary et~al.(2014)Ghifary, Kleijn, and Zhang]{ae_adaptation}
Muhammad Ghifary, W~Bastiaan Kleijn, and Mengjie Zhang.
\newblock Domain adaptive neural networks for object recognition.
\newblock In \emph{PRICAI: Trends in Artificial Intelligence}, pp.\  898--904.
  2014.

\bibitem[Gong et~al.(2012)Gong, Shi, Sha, and Grauman]{gfk}
Boqing Gong, Yuan Shi, Fei Sha, and Kristen Grauman.
\newblock Geodesic flow kernel for unsupervised domain adaptation.
\newblock In \emph{CVPR}, pp.\  2066--2073, 2012.

\bibitem[Gong et~al.(2013)Gong, Grauman, and Sha]{landmark}
Boqing Gong, Kristen Grauman, and Fei Sha.
\newblock Connecting the dots with landmarks: Discriminatively learning
  domain-invariant features for unsupervised domain adaptation.
\newblock In \emph{ICML}, pp.\  222--230, 2013.

\bibitem[Gopalan et~al.(2011)Gopalan, Li, and Chellappa]{gopalan2011domain}
Raghuraman Gopalan, Ruonan Li, and Rama Chellappa.
\newblock Domain adaptation for object recognition: An unsupervised approach.
\newblock In \emph{ICCV}, pp.\  999--1006, 2011.

\bibitem[Gretton et~al.(2012)Gretton, Borgwardt, Rasch, Sch{\"o}lkopf, and
  Smola]{mmd}
Arthur Gretton, Karsten~M Borgwardt, Malte~J Rasch, Bernhard Sch{\"o}lkopf, and
  Alexander Smola.
\newblock A kernel two-sample test.
\newblock \emph{The Journal of Machine Learning Research}, 13\penalty0
  (1):\penalty0 723--773, 2012.

\bibitem[Griffin et~al.(2007)Griffin, Holub, and Perona]{griffin2007caltech}
Gregory Griffin, Alex Holub, and Pietro Perona.
\newblock Caltech-256 object category dataset.
\newblock 2007.

\bibitem[He et~al.(2016)He, Zhang, Ren, and Sun]{resnet}
Kaiming He, Xiangyu Zhang, Shaoqing Ren, and Jian Sun.
\newblock Deep residual learning for image recognition.
\newblock \emph{CVPR}, 2016.

\bibitem[Huang et~al.(2006)Huang, Gretton, Borgwardt, Sch{\"o}lkopf, and
  Smola]{huang2006correcting}
Jiayuan Huang, Arthur Gretton, Karsten~M Borgwardt, Bernhard Sch{\"o}lkopf, and
  Alex~J Smola.
\newblock Correcting sample selection bias by unlabeled data.
\newblock In \emph{NIPS}, pp.\  601--608, 2006.

\bibitem[Ioffe \& Szegedy(2015)Ioffe and Szegedy]{bn}
Sergey Ioffe and Christian Szegedy.
\newblock Batch normalization: Accelerating deep network training by reducing
  internal covariate shift.
\newblock In \emph{ICML}, pp.\  448--456, 2015.

\bibitem[Jia et~al.(2014)Jia, Shelhamer, Donahue, Karayev, Long, Girshick,
  Guadarrama, and Darrell]{caffe}
Yangqing Jia, Evan Shelhamer, Jeff Donahue, Sergey Karayev, Jonathan Long, Ross
  Girshick, Sergio Guadarrama, and Trevor Darrell.
\newblock Caffe: Convolutional architecture for fast feature embedding.
\newblock In \emph{ACM MM}, pp.\  675--678, 2014.

\bibitem[Khosla et~al.(2012)Khosla, Zhou, Malisiewicz, Efros, and
  Torralba]{khosla2012undoing}
Aditya Khosla, Tinghui Zhou, Tomasz Malisiewicz, Alexei~A Efros, and Antonio
  Torralba.
\newblock Undoing the damage of dataset bias.
\newblock In \emph{ECCV}, pp.\  158--171. 2012.

\bibitem[Krizhevsky et~al.(2012)Krizhevsky, Sutskever, and Hinton]{alexnet}
Alex Krizhevsky, Ilya Sutskever, and Geoffrey~E Hinton.
\newblock Imagenet classification with deep convolutional neural networks.
\newblock In \emph{NIPS}, pp.\  1097--1105, 2012.

\bibitem[Long et~al.(2015)Long, Cao, Wang, and Jordan]{dan}
Mingsheng Long, Yue Cao, Jianmin Wang, and Michael Jordan.
\newblock Learning transferable features with deep adaptation networks.
\newblock In \emph{ICML}, pp.\  97--105, 2015.

\bibitem[Long et~al.(2016)Long, Wang, and Jordan]{long2016unsupervised}
Mingsheng Long, Jianmin Wang, and Michael~I Jordan.
\newblock Unsupervised domain adaptation with residual transfer networks.
\newblock In \emph{NIPS}, 2016.

\bibitem[Pan et~al.(2011)Pan, Tsang, Kwok, and Yang]{tca}
Sinno~Jialin Pan, Ivor~W Tsang, James~T Kwok, and Qiang Yang.
\newblock Domain adaptation via transfer component analysis.
\newblock \emph{IEEE Transactions on Neural Networks}, 22\penalty0
  (2):\penalty0 199--210, 2011.

\bibitem[Patel et~al.(2015)Patel, Gopalan, Li, and Chellappa]{patel2015visual}
Vishal~M Patel, Raghuraman Gopalan, Ruonan Li, and Rama Chellappa.
\newblock Visual domain adaptation: A survey of recent advances.
\newblock \emph{IEEE Signal Processing Magazine}, 32\penalty0 (3):\penalty0
  53--69, 2015.

\bibitem[Russakovsky et~al.(2015)Russakovsky, Deng, Su, Krause, Satheesh, Ma,
  Huang, Karpathy, Khosla, Bernstein, et~al.]{imagenet}
Olga Russakovsky, Jia Deng, Hao Su, Jonathan Krause, Sanjeev Satheesh, Sean Ma,
  Zhiheng Huang, Andrej Karpathy, Aditya Khosla, Michael Bernstein, et~al.
\newblock {ImageNet} large scale visual recognition challenge.
\newblock \emph{International Journal of Computer Vision}, 115\penalty0
  (3):\penalty0 211--252, 2015.

\bibitem[Saenko et~al.(2010)Saenko, Kulis, Fritz, and Darrell]{office}
Kate Saenko, Brian Kulis, Mario Fritz, and Trevor Darrell.
\newblock Adapting visual category models to new domains.
\newblock In \emph{ECCV}, pp.\  213--226. 2010.

\bibitem[Shimodaira(2000)]{shimodaira2000improving}
Hidetoshi Shimodaira.
\newblock Improving predictive inference under covariate shift by weighting the
  log-likelihood function.
\newblock \emph{Journal of statistical planning and inference}, 90\penalty0
  (2):\penalty0 227--244, 2000.

\bibitem[Sun \& Saenko(2016)Sun and Saenko]{sun2016deep}
Baochen Sun and Kate Saenko.
\newblock Deep coral: Correlation alignment for deep domain adaptation.
\newblock \emph{arXiv preprint arXiv:1607.01719}, 2016.

\bibitem[Sun et~al.(2016)Sun, Feng, and Saenko]{coral}
Baochen Sun, Jiashi Feng, and Kate Saenko.
\newblock Return of frustratingly easy domain adaptation.
\newblock \emph{AAAI}, 2016.

\bibitem[Szegedy et~al.(2015)Szegedy, Vanhoucke, Ioffe, Shlens, and
  Wojna]{inception_v3}
Christian Szegedy, Vincent Vanhoucke, Sergey Ioffe, Jonathon Shlens, and
  Zbigniew Wojna.
\newblock Rethinking the inception architecture for computer vision.
\newblock \emph{arXiv preprint arXiv:1512.00567}, 2015.

\bibitem[Tommasi et~al.(2015)Tommasi, Patricia, Caputo, and
  Tuytelaars]{deeper_bias}
Tatiana Tommasi, Novi Patricia, Barbara Caputo, and Tinne Tuytelaars.
\newblock A deeper look at dataset bias.
\newblock \emph{German Conference on Pattern Recognition}, 2015.

\bibitem[Torralba \& Efros(2011)Torralba and Efros]{unbiased}
Antonio Torralba and Alexei~A Efros.
\newblock Unbiased look at dataset bias.
\newblock In \emph{CVPR}, pp.\  1521--1528, 2011.

\bibitem[Tzeng et~al.(2014)Tzeng, Hoffman, Zhang, Saenko, and Darrell]{ddc}
Eric Tzeng, Judy Hoffman, Ning Zhang, Kate Saenko, and Trevor Darrell.
\newblock Deep domain confusion: Maximizing for domain invariance.
\newblock \emph{arXiv preprint arXiv:1412.3474}, 2014.

\bibitem[Tzeng et~al.(2015)Tzeng, Hoffman, Darrell, and Saenko]{joint}
Eric Tzeng, Judy Hoffman, Trevor Darrell, and Kate Saenko.
\newblock Simultaneous deep transfer across domains and tasks.
\newblock In \emph{ICCV}, pp.\  4068--4076, 2015.

\bibitem[Van~der Maaten \& Hinton(2008)Van~der Maaten and Hinton]{tsne}
Laurens Van~der Maaten and Geoffrey Hinton.
\newblock Visualizing data using t-sne.
\newblock \emph{Journal of Machine Learning Research}, 9\penalty0
  (2579-2605):\penalty0 85, 2008.

\bibitem[Yosinski et~al.(2014)Yosinski, Clune, Bengio, and
  Lipson]{transferable}
Jason Yosinski, Jeff Clune, Yoshua Bengio, and Hod Lipson.
\newblock How transferable are features in deep neural networks?
\newblock In \emph{NIPS}, pp.\  3320--3328, 2014.

\end{thebibliography}
\bibliographystyle{iclr2017_conference}
}

\end{document}